%% file: arxiv.tex
\documentclass[10pt,twocolumn,letterpaper]{article}

\usepackage[pagenumbers]{cvpr} 

\usepackage{graphicx}
\usepackage{amsmath}
\usepackage{amssymb}
\usepackage{booktabs}

\usepackage[pagebackref,breaklinks,colorlinks]{hyperref}

\usepackage[capitalize]{cleveref}
\crefname{section}{Sec.}{Secs.}
\Crefname{section}{Section}{Sections}
\Crefname{table}{Table}{Tables}
\crefname{table}{Tab.}{Tabs.}

\input{commands}

\makeatletter
\renewcommand\paragraph{\@startsection{paragraph}{4}{\z@}%
                                    {1.25ex \@plus1ex \@minus.2ex}%
                                    {-1em}%
                                    {\normalfont\normalsize\bfseries}}
\makeatother

\begin{document}

\title{\titletext}


\author{
Alexandre Boulch$^{1}$ \quad
Corentin Sautier$^{1,2}$ \quad
Bj\"orn Michele$^{1,3}$ \quad
Gilles Puy$^{1}$ \quad
Renaud Marlet$^{1,2}$
\and \quad
\and
\large
\hspace{-3mm}\textsuperscript{1}Valeo.ai, Paris, France  \hspace{1mm} \textsuperscript{2}LIGM, Ecole des Ponts, Univ Gustave Eiffel, CNRS, Marne-la-Vall\'ee, France \\
\textsuperscript{3}CNRS, IRISA, Univ. Bretagne Sud, Vannes, France
}

\maketitle

\input{sections/abstract}


\input{main}

{\small
\bibliographystyle{ieee_fullname}
\bibliography{egbib}
}

\clearpage
\appendix
\input{sections/appendix}
\end{document}

%% file: commands.tex
\usepackage[utf8]{inputenc} 
\usepackage[T1]{fontenc}    
\usepackage{hyperref}       
\usepackage{url}            
\usepackage{booktabs}       
\usepackage{amsfonts}       
\usepackage{nicefrac}       
\usepackage{microtype}      
\usepackage[dvipsnames]{xcolor}         

\usepackage{amssymb}
\usepackage{pifont}
\usepackage{fixmetodonotes}
\usepackage{colortbl}
\usepackage{multirow}
\usepackage{ulem}

\usepackage{listings}
\usepackage{xcolor}

\usepackage{minitoc}

\definecolor{codegreen}{rgb}{0,0.6,0}
\definecolor{codegray}{rgb}{0.5,0.5,0.5}
\definecolor{codepurple}{rgb}{0.58,0,0.82}
\definecolor{backcolour}{rgb}{0.95,0.95,0.92}

\lstdefinestyle{mystyle}{
    backgroundcolor=\color{backcolour},   
    commentstyle=\color{codegreen},
    keywordstyle=\color{magenta},
    numberstyle=\tiny\color{codegray},
    stringstyle=\color{codepurple},
    basicstyle=\ttfamily\footnotesize,
    breakatwhitespace=false,         
    breaklines=true,                 
    captionpos=b,                    
    keepspaces=true,                 
    numbers=left,                    
    numbersep=5pt,                  
    showspaces=false,                
    showstringspaces=false,
    showtabs=false,                  
    tabsize=2,
}

\definecolor{MyGreen}{RGB}{0, 180, 0}
\definecolor{MyRed}{RGB}{180, 0, 0}
\definecolor{MyBlue}{RGB}{30, 0, 180}

\newcommand{\cmark}{{\textcolor{MyGreen}{\ding{51}}}}%
\newcommand{\xmark}{{\textcolor{MyRed}{\ding{55}}}}%

\newcommand{\tplus}[1]{\textcolor{MyGreen}{#1}}
\newcommand{\tminus}[1]{\textcolor{MyRed}{#1}}

\newcommand{\method}{{ALSO}}

\newcommand{\titletext}{
ALSO: Automotive Lidar Self-supervision by Occupancy estimation
}

\newcommand{\rthree}[1]{{#1}}
\newcommand{\rtwo}[1]{\cellcolor{MyBlue!10}{#1}}
\newcommand{\rone}[1]{\cellcolor{MyBlue!30}{#1}}

\newcommand{\pone}[1]{\cellcolor{MyGreen!5}{#1}}
\newcommand{\ptwo}[1]{\cellcolor{MyGreen!15}{#1}}
\newcommand{\pthree}[1]{\cellcolor{MyGreen!25}{#1}}
\newcommand{\pfour}[1]{\cellcolor{MyGreen!35}{#1}}
\newcommand{\pfive}[1]{\cellcolor{MyGreen!45}{#1}}
\newcommand{\psix}[1]{\cellcolor{MyGreen!55}{#1}}

\newcommand{\mone}[1]{\cellcolor{MyRed!5}{#1}}
\newcommand{\mtwo}[1]{\cellcolor{MyRed!15}{#1}}
\newcommand{\mthree}[1]{\cellcolor{MyRed!25}{#1}}

\newcommand{\pointcloud}{P}
\newcommand{\point}{p}

\newcommand{\querycloud}{Q}
\newcommand{\query}{q}

\newcommand{\ifront}{\mathsf{front}}
\newcommand{\ibehind}{\mathsf{behind}}
\newcommand{\isight}{\mathsf{sight}}
\newcommand{\vlos}{u}

\newcommand{\intensity}{i}
\newcommand{\intensityest}{\hat{\intensity}}

\newcommand{\supports}{S}
\newcommand{\support}{s}

\newcommand{\latent}{z}

\newcommand{\occupancy}{o}
\newcommand{\occupancyest}{\hat{\occupancy}}

\newcommand{\sensor}{c}

\newcommand{\loss}{\mathcal{L}}
\newcommand{\lossrec}{\mathcal{L}_{\mathsf{occup}}}
\newcommand{\lossint}{\mathcal{L}_{\mathsf{intens}}}

\newcommand{\renaud}[1]{{#1}}

\newcommand{\alex}[1]{{#1}}

\newcommand{\gilles}[1]{{#1}}

\newcommand{\greyrule}[1]{\arrayrulecolor{black!20} #1\arrayrulecolor{black}}

%% file: sections/abstract.tex
\begin{abstract}

We propose a new self-supervised method for pre-training the backbone of deep perception models operating on point clouds. The core idea is to train the model on a pretext task which is the reconstruction of the surface on which the 3D points are sampled, and to use the underlying latent vectors as input to the perception head. The intuition is that if the network is able to reconstruct the scene surface, given only sparse input points, then it probably also captures some fragments of semantic information, that can be used to boost an actual perception task. This principle has a very simple formulation, which makes it both easy to implement and widely applicable to a large range of 3D sensors and deep networks performing semantic segmentation or object detection. In fact, it supports a single-stream pipeline, as opposed to most contrastive learning approaches, allowing training on limited resources. We conducted extensive experiments on various autonomous driving datasets, involving very different kinds of lidars, for both semantic segmentation and object detection. The results show the effectiveness of our method to learn useful representations without any annotation, compared to existing approaches.

The code is available at \href{https://github.com/valeoai/ALSO}{github.com/valeoai/ALSO}

\end{abstract}

%% file: main.tex
\input{sections/introduction}
\input{sections/related}
\input{sections/method}
\input{sections/experiments}

\input{sections/conclusion}

%% file: sections/introduction.tex
\input{figures/occupancy.tex}

\input{figures/overview.tex}

\section{Introduction}

As a complement to 2D cameras, lidars 
\renaud{directly} capture the 3D environment of a vehicle with high accuracy and low sensitivity to adverse conditions, such as low illumination\renaud{, bright sunlight or oncoming headlights}. 
They are thus essential sensors for safe autonomous driving.

Most state-of-the-art lidar-based perception methods, whether they regard semantic segmentation \cite{cylinder3d,spvcnn,minkowskicnn} or object detection \cite{second,pvrcnn,centerpoint,pointpillar}, assume they can be trained on large annotated datasets. However, annotating 3D data for such tasks is notoriously costly and time consuming. As data acquisition is much cheaper than data annotation, being able to leverage unannotated data to increase the performance or reduce the annotation effort is a significant asset.

A promising direction to address this question is to pre-train a neural network using only unannotated data, e.g., on a pretext task which does not require manual labelling, and then to fine-tune the resulting self-supervised pre-trained network for the targeted downstream task(s). With adequate pre-training, the learned network weights are a good starting point for further supervised optimization; training a specific downstream task then typically requires fewer annotations to reach the same performance level as if trained from scratch.

A number of self-supervised approaches have been very successful in 2D (images), even reaching the level of supervised pre-training \cite{chen2020mocov2,caron2021dino,byol,he2022masked}.
Some self-supervised ideas have been proposed for 3D data as well, which are  often transpositions in 3D of 2D methods~\cite{strl,voxelmae}. Most of them focus on contrastive learning \cite{segcontrast,pointcontrast,slidr,depthcontrast,proposalcontrast} which learns to infer perceptual features that are analogous for similar objects while being far apart for dissimilar objects.

Only few such methods apply to lidar point clouds, which have the particularity of having very heterogeneous densities.

In this work, we propose a totally new pretext task for the self-supervised pre-training of neural networks operating on point clouds. We observe that one of the main reasons why downstream tasks may fail is related to the sparsity of data. Indeed, with automotive lidars, 3D points are especially sparse when far from the sensor or on areas where laser beams have a high incidence on the scanned surface. In such cases, objects are difficult to recognize, and even more so if they are small, such as so-called vulnerable road user (e.g., pedestrians, bicyclists) and traffic signs.

\alex{In a \renaud{mostly} supervised context, geometric information such as object shape
~\cite{kundu20183d,najibi2020dops} and visibility information~\cite{hu2020you} have proved to boost detection performance.}
Our approach uses \renaud{visibility-based} surface reconstruction as a pretext task for self-supervision. It takes root in the implicit shape representation literature, where shapes are encoded into latent vectors, that can be decoded into a function indicating the shape volume occupancy or the distance to the shape surface. 
The intuitive idea is that if a network is able to properly reconstruct the 3D geometry of a scene from point clouds, then there are good chances that it constructs rich features that can be reused in a number of other contexts, in particular regarding semantic-related tasks.


\alex{Our contributions are as follows: (1)~we 
combine surface reconstruction
and visibility \renaud{information} to create a \renaud{sensor-agnostic} and \renaud{backbone-agnostic} pretext task on 3D point clouds, 
which produces good \renaud{self-supervised} point features for semantic segmentation and object detection; (2)~we design a loss that leads each 
point to capture enough knowledge to reconstruct its neighborhood (instead of aggregating 
\gilles{information from neighbors} for a more accurate surface reconstruction), which instils \renaud{a taste of} semantics in
the geometric task; 
}
\renaud{(3)~based on 
experiments across seven datasets, our self-supervised features, that require only limited resources for training (single 16G GPU), outperform state-of-the-art self-supervised features on semantic segmentation, and are on par with these features on object detection.} 

%% file: figures/occupancy.tex



\begin{figure}
    \small
    \centering
    \setlength{\tabcolsep}{1pt}
    \begin{tabular}{cc}
        \includegraphics[width=0.49\linewidth]{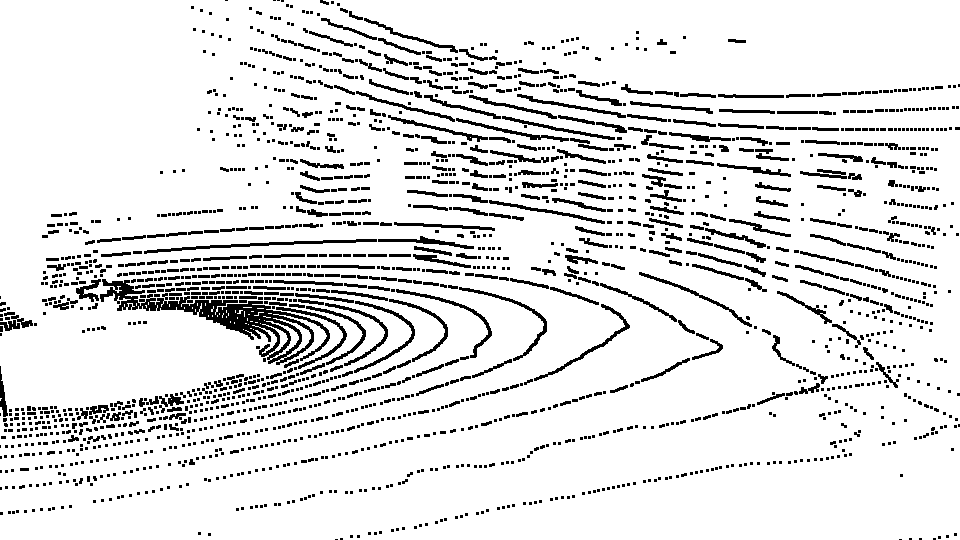} & 
        \includegraphics[width=0.49\linewidth]{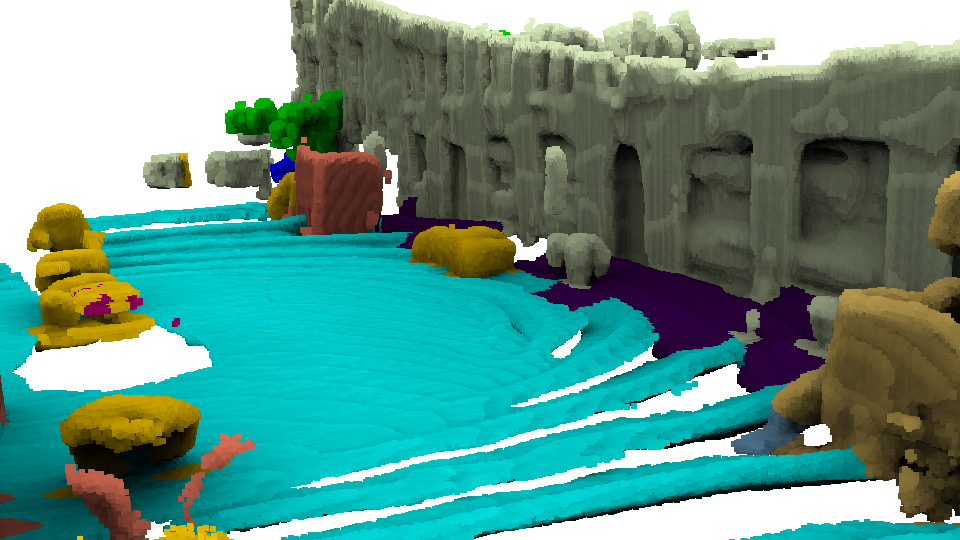} \\
        \includegraphics[width=0.49\linewidth]{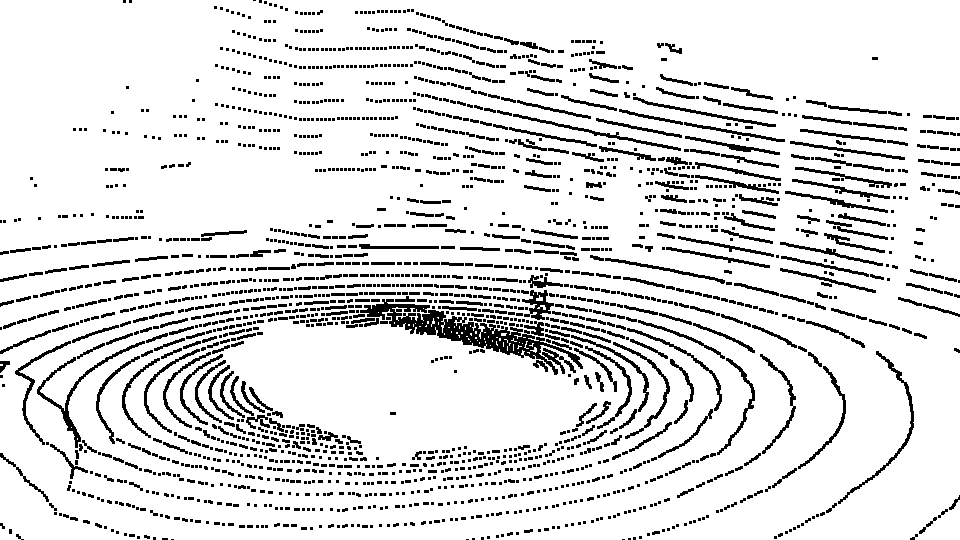} & 
        \includegraphics[width=0.49\linewidth]{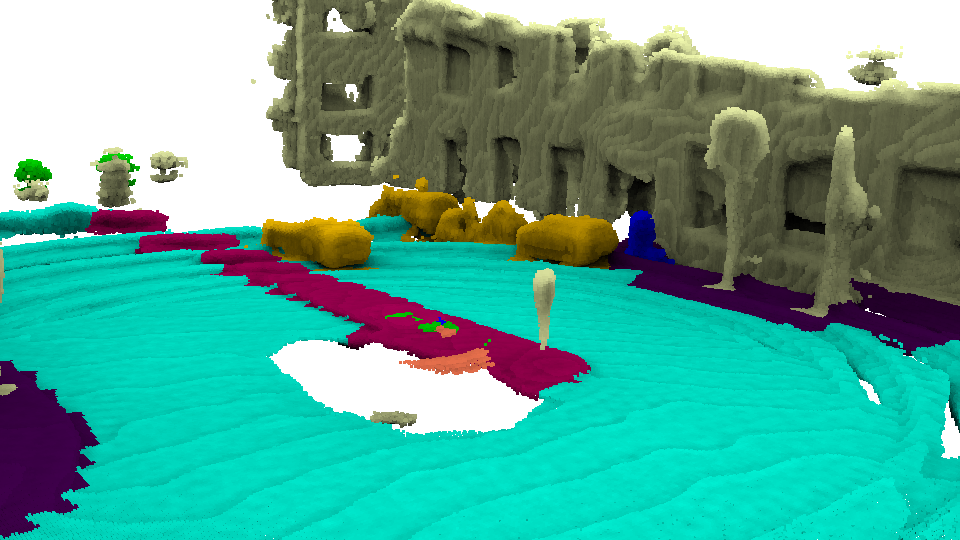} \\
        \multicolumn{2}{c}{(a) nuScenes.} \\
        \includegraphics[width=0.49\linewidth]{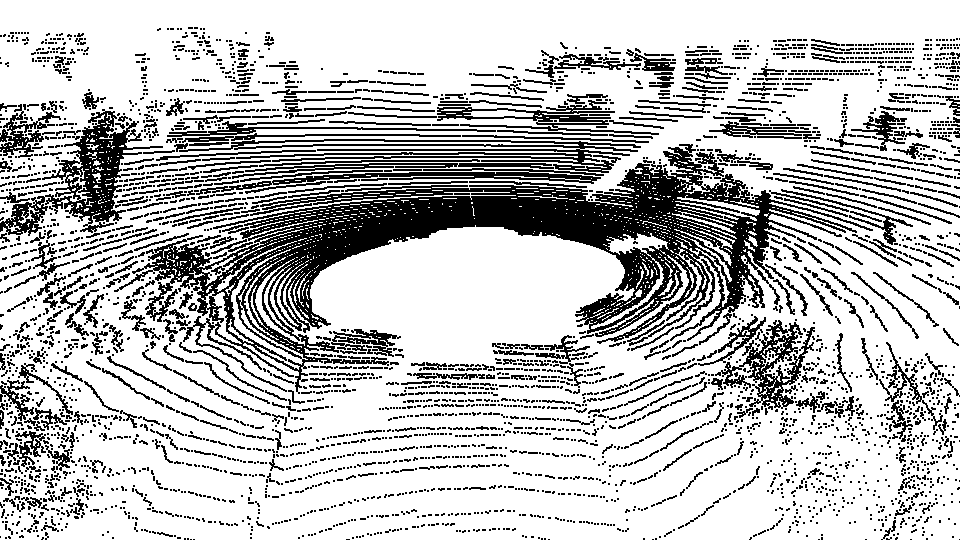} &
        \includegraphics[width=0.49\linewidth]{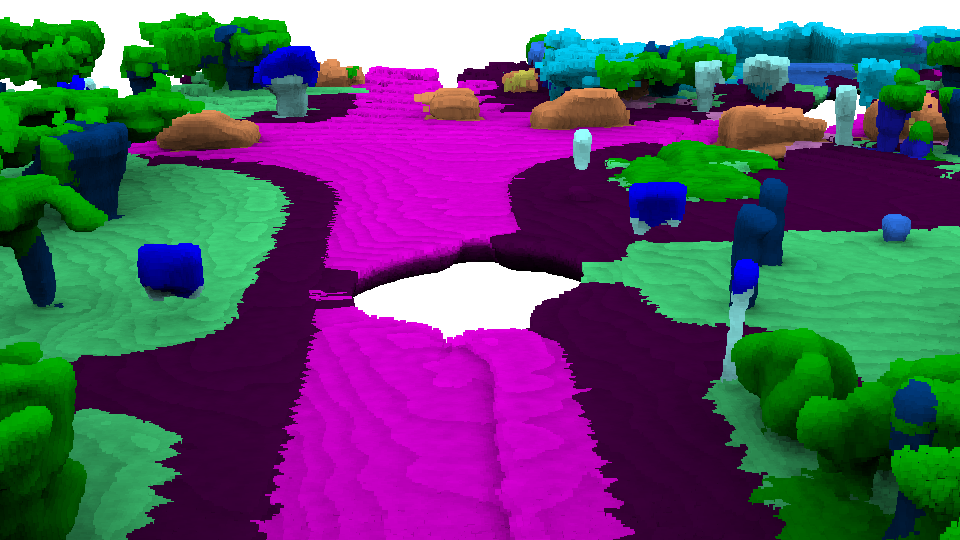} \\
        \includegraphics[width=0.49\linewidth]{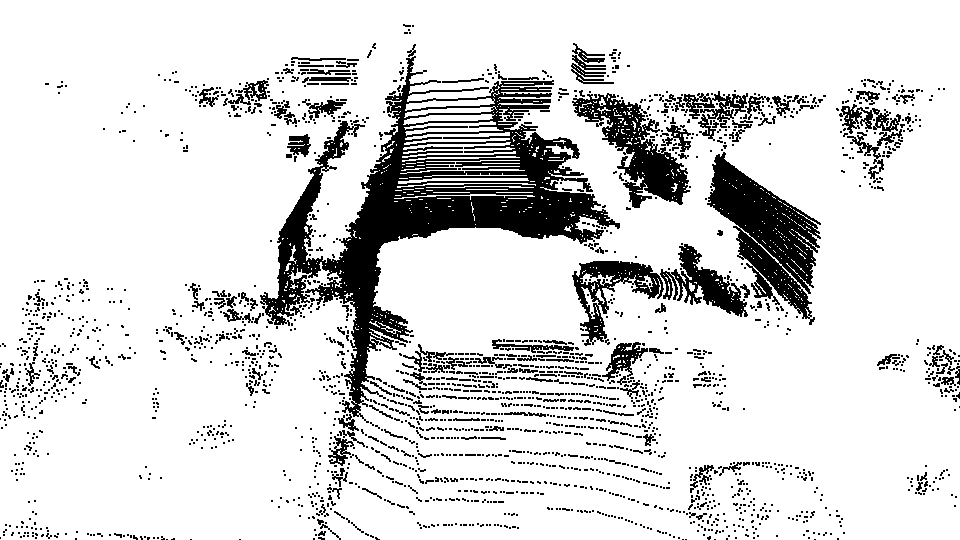} &
        \includegraphics[width=0.49\linewidth]{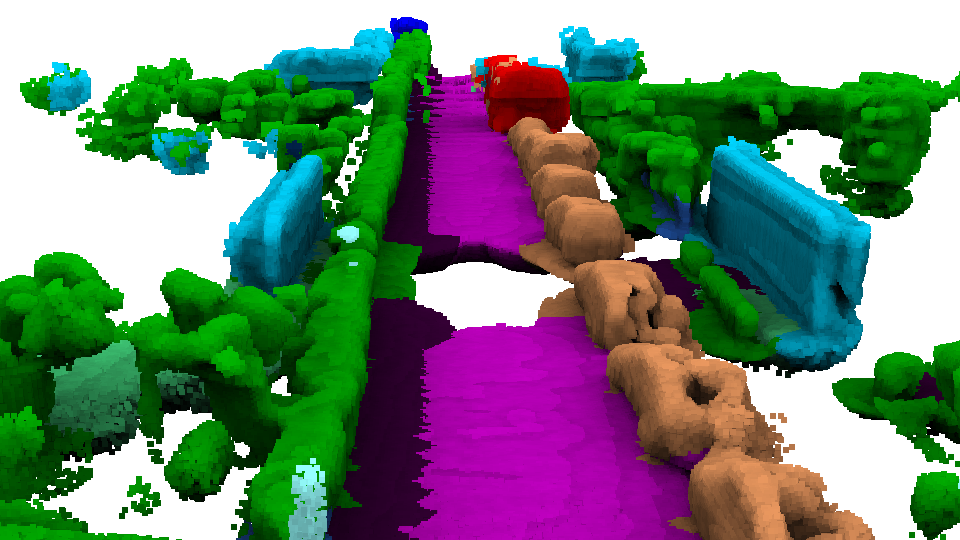} \\
        \multicolumn{2}{c}{(b) SemanticKITTI.} \\
    \end{tabular}
    
    \caption{Aggregation of the self-supervised training on lidar datasets. Input point cloud (first column) and occupancy prediction colored by the learned downstream labels.}
    \label{fig:qualitative_occ_seg}
\end{figure}

%% file: figures/overview.tex
\begin{figure*}
  \centering
  \includegraphics[width=\linewidth]{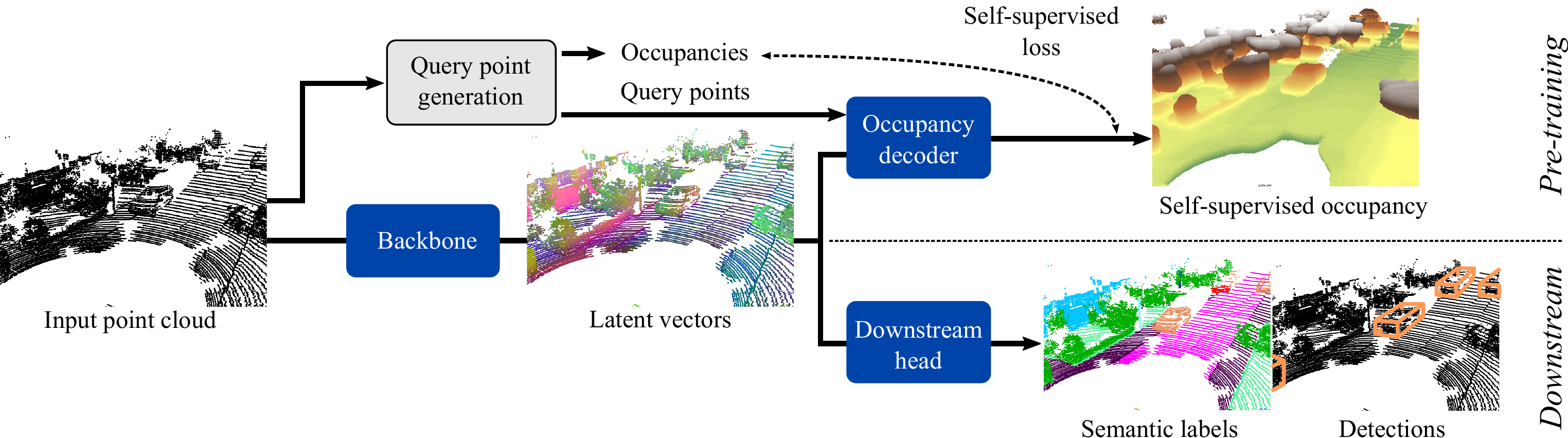}
  \caption{Overview of the approach. The backbone to pre-train produces latent vectors for each input point. At pre-training time, the latent vector are fed into an volumetric occupancy head that classifies query points as full or empty. At semantic training or test time, the same latent vectors are fed into a semantic head, e.g., for semantic segmentation or object detection.}
  \label{fig:overview}
  \vspace{-1mm}
\end{figure*}

%% file: sections/related.tex
\section{Related work}
\label{sec:related}

Self-supervision has been the subject of a significant research effort, including for image applications. Early methods mostly focused on direct estimation of the transformation applied to images \cite{gidaris2018unsupervised,zhang2016colorful,doersch2015unsupervised,misra2016shuffle,noroozi2016unsupervised}. More recently a great boost of performances has been accomplished with contrastive learning \cite{chen2020simple,misra2020self,tian2020contrastive,wu2018unsupervised}, clustering based methods~\cite{deepcluster,swav} and reconstruction-based approaches. The latter can operate the reconstruction in the feature space \cite{bownet,obow,byol,caron2021dino,chen2020mocov2} trying to reconstruct the features issued from a teacher signal or in the images domain \cite{he2022masked,bao2021beit}, a partially masked input image being reconstructed.

\subsection{2D adaptation to point clouds} 

2D methods have being adapted to point cloud, in particular for detection.
In the dataset presentation paper of ONCE~\cite{once}, the authors produce self-supervision baselines using methods adapted from image self-supervision: BYOL~\cite{byol}, SWaV~\cite{swav} and DeepCluster~\cite{deepcluster}. Another work, related to MAE~\cite{he2022masked} (images) or Point-MAE~\cite{pointmae} (part segmentation): Voxel-MAE~\cite{voxelmae} reconstruct the complete voxel grid, given a partially masked input.

\textit{Our approach} is a reconstruction approach. The major difference lies in our supervision signal which is not made by masking an already sparse input, but by estimating the unknown underlying scene surface using sensor information.

\subsection{Self-supervision for point clouds.}

\noindent\textbf{Classification and part segmentation.}
Following the same trends as image methods, pretext tasks have been built in order to reconstruct the input point cloud \cite{sauder2019self,Wang2020PreTrainingBC}, estimate a global transformation \cite{chen2021shape,Poursaeed2020SelfSupervisedLO}, contrast between objects views \cite{sanghi2020info3d,wang2021unsupervised,chen2021unsupervised,du2021self} or estimate clusters \cite{Hassani2019UnsupervisedMF,zhang_3dv_2019}.

\noindent\textbf{Semantic segmentation.}
Scene level pre-training has been tackled using multi-temporal data, for example, by contrasting point-wise representation which are matched across two temporally distinct and registered acquisitions of the same scene \cite{pointcontrast, Hou_2021_CVPR,coconets}. Segcontrast \cite{segcontrast} extract segments likely to belong to the same object (ground plane using RANSAC \cite{fischler1981random}, other cluster using DBSCAN \cite{ester1996density}) and then contrast between segment representations for different augmentation of the same scene. DepthContrast \cite{depthcontrast} contrast four representations obtained with two networks and two augmented views of the same scene. The method is shown to be efficient for indoor data as well as outdoor data, but requires a joint-training of two networks, thus is memory intensive. In STRL \cite{strl}, a model and an exponential moving average version of it are fed two temporally-close point cloud frames, altered with various 3D augmentations, and the objective is that both representation are similar.

Another class of methods explores cross-modality, leveraging one or multiple images \cite{slidr,ppkt,simipu}. In our case, we condider that only lidar modality is available.

\textit{Our method} is, as opposed to all previously cited approaches, not a contrastive method. We do not rely on several augmentations of the scenes, processed in parallel to build our representations. Therefore, our approach can be easily trained on a single 16GB memory GPU.

\noindent\textbf{Detection.}
Several methods have been adapted from semantic segmentation to detection. PointContrast \cite{pointcontrast}, DepthContrast \cite{depthcontrast} or STRL~\cite{strl} propose an extension to detection by training the 3D backbone of the detectors \cite{second,pvrcnn}.

The current best performing methods are specially designed for 3D detection. GCC3D \cite{gcc3d} uses both a contrastive and clustering mechanisms to learn a 3D object detection encoder. Augmented versions of a scene are encoded and a local contrastive loss is applied to enforce feature invariance. Feature learning is then refined with a clustering objective using temporal clues. ProposalContrast \cite{proposalcontrast} applies contrastive learning at region level arguing that scene-level representation may lose details while point-level contrast favors small receptive field, without object-level knowledge.

\textit{Our approach} also achieves object-size knowledge, but we only require a single parameter, which is the size of the neighborhood to be reconstructed, which intuitively should have a similar dimension to objects in the scene. Moreover, our method can be applied indifferently for semantic segmentation or detection, contrarily to the best performing methods \cite{gcc3d,proposalcontrast,voxelmae}.

\subsection{Occupancy reconstruction}

Surface reconstruction is a well studied subject in computational geometry.
Surfaces are usually described either using explicit representations (voxels \cite{Wu2015CVPR, Maturana2015IROS}, surface point clouds \cite{Achlioptas2018ICML, Lin2018AAAI, Yang2019ICCV} or meshes \cite{Nash2020polygen,Groueix2018CVPR, Wang2018ECCV,  Liao2018CVPR, Gkioxari2019ICCV, Luo2021AAAI}) or with implicit representations, which define a function over the 3D space from which the surface can be extracted.

\noindent\textbf{Implicit reconstruction with deep networks.}
Implicit representations have gained in popularity since the seminal work DeepSDF \cite{deepsdf}. The existing methods estimate either a distance function \cite{deepsdf, Michalkiewicz2019ICCV, Gropp2020ICML,Chibane2020Neural} (signed or not), an occupancy function \cite{Chen2019CVPR, Mescheder2019CVPR,poco} or both \cite{Erler2020Points2Surf}.

\textit{Our approach} predicts an occupancy function, i.e., label the 3D space as inside or outside the volume.

\noindent\textbf{Global and local representations.}
Surface reconstruction from point clouds has mainly been tackled with two distinct objectives.
On the one hand, shape representation \cite{deepsdf,Mescheder2019CVPR,Cai2020ShapeGF} aims at associating each object with a single latent vector containing rich global geometric information, in a space suitable for interpolation, possibly with good classification properties \cite{boulch2021needrop}. On the other hand, surface reconstruction using local representation \cite{Jiang2020CVPR, Chibane2020CVPR, Chibane2020Neural, convonet, poco, Erler2020Points2Surf, williams2022neural} aims at precise surface estimation and is able to scale to large scenes. However, the local description focuses on low-level geometric information, rather than object-level knowledge needed for self-supervision.

\textit{Our method} intends to exploit the properties of both categories.
As large outdoor scenes are composed of multiple objects and surfaces, we propose to learn the occupancy using a local approach, but with the objective of learning object-level representations.

\noindent\textbf{Supervision.}
Thanks to the synthetic datasets such as ModelNet~\cite{modelnet}, ShapeNet~\cite{shapenet} or ABC~\cite{abc}, it is possible to obtain an occupancy ground truth used for supervision. Local approaches supervised on these shapes are able to generalize to scene level when provided an additional orientation information such as point normals \cite{Jiang2020CVPR, poco} or sensor location \cite{Raphael}.

Other methods have been developed to perform volume reconstruction without the need for any ground truth label, focusing on loss design. Sign agnostic losses are used in SAL \cite{Atzmon2020CVPR} and SALD \cite{Atzmon2021ICLR}. It minimizes the norm of the estimated signed distance field with respect to input-computed distance field. A careful initialization of the network is needed to ensure a signed output. IGR \cite{Gropp2020ICML} encourages a null distance at each input point, while enforcing a non-null gradient norm, thus favoring sign changes. Needrop \cite{boulch2021needrop} uses a loss applied on a segment sampled such that the middle point is an input point and force both extremities to have opposite labels.

\textit{Our approach}, in contrast, does not rely on the design of a specific loss.
We use the sensor information as in \cite{Raphael} to generate points where we can estimate the occupancy with confidence. We can then train our model with a binary cross-entropy as if in a supervised setting.

%% file: sections/method.tex
\section{Method}

We propose surface reconstruction as a pretext task to create self-supervised features for 3D point clouds, that are well suited for semantic segmentation and object detection.

As mentioned in Section~\ref{sec:related}, networks can be trained with supervision to estimate implicitly the surface underlying a given point cloud. Besides, using visibility information has been shown an effective way to improve surface reconstruction by adding extra points to supervise the training \cite{Raphael}. Inspired by these works, we propose to reuse rich shape features for downstream tasks. To create such features without the need for manual annotations, we propose to use visibility information for unsupervised surface reconstruction. What's more, we adapt surface reconstruction to produce intermediate latent vectors that capture not only geometrical details but also some semantic knowledge. The overall principle of the method is presented in Figure~\ref{fig:overview}.

\subsection{Support points and latent vectors}

In methods such as POCO \cite{poco}, resp.\ ConvONet \cite{convonet}, a latent vector is first computed for each input point, resp.\ each pre-defined (2D or 3D) grid point. The occupancy of a given query point is then predicted from the latent vectors of neighboring points \cite{poco}, resp.\ neighboring grid points \cite{convonet}.

Similarly, in our setting, we consider that input points $\point \in \pointcloud$, with optional intensity $\intensity_\point$, are given to a backbone that outputs, on given support points $\support \in \supports$, an associated latent vector $\latent_\support$. For semantic segmentation, the support points are the input points. For object detection, using detectors such as SECOND~\cite{second} or PVRCNN~\cite{pvrcnn}, the support points are 2D points on a grid in the bird-eye-view (BEV) plane.

\input{figures/queries.tex}

\subsection{Generating query points for self-supervision}
\label{sec:query_points}

To introduce self-supervision, we create query points $\query \in \querycloud$ with known occupancy $\occupancy_\query$, although no ground truth surface is used or even available. To that end, we exploit visibility information, knowing the sensor location.

Given a 3D point $\point$ sampled on the surface of the scene by a sensor whose center is located at $\sensor$, we consider that points \emph{in front of} $\point$, i.e., on the 3D segment $[\sensor,\point]$, are empty, while points immediately \emph{behind} the observed point $\point$ along the line of sight of the sensor are not (see Figure~\ref{fig:queries}(a)).

Concretely, as in \cite{Raphael}, for each input point $\point$, we create two query points $\query_\ifront = \point - \delta_\point$ and $\query_\ibehind = \point + \delta_\point$, where $\delta_\point = \delta \vlos$, $\delta > 0$ is a small distance, and $\vlos = (\sensor-p)/\lVert \sensor-p \rVert$ is a unit vector pointing from the sensor location $\sensor$ to the observed point $\point$. $\query_\ifront$ is considered empty and $\query_\ibehind$ full.
While $\query_\ifront$ is empty for sure (unless $p$ is an outlier), $\query_\ibehind$ is not necessarily full in case the object is very thin (thinner than $\delta$) or at the border of objects for grazing lines of sight. Nevertheless, \cite{Raphael} shows that this hypothesis is correct enough in general, leading to significant benefits in surface reconstruction. The parameter $\delta$ however has to be adjusted to the expected minimum thickness of scene objects.

Additionally, we create a third empty query point $\query_\isight$ randomly picked in the segment $[\sensor,\point]$. These query points $\query \in \querycloud$ and associated occupancies $\occupancy_\query$ are used as the supervision signal to pre-train the backbone.

\subsection{Estimating occupancy towards semantization}

In surface reconstruction methods, the goal is to infer the most accurate reconstruction. To that end, these methods rely on different forms of interpolation \cite{convonet,poco}, gathering information from latent vectors of neighboring support points. 

Here, our goal is different. We do not care so much about geometrical details. What we want is to infer features that are typical to the underlying objects or object parts. To that end, we reverse the reconstruction paradigm: instead of estimating occupancy by combining fine local information from neighbors, we encourage features of neighbors to be similar \renaud{by enforcing the feature of a point to be able to reconstruct a whole ball around it}. Surface reconstruction is then \renaud{possibly} less accurate, but the inferred features are more global to the object or object part, which makes semantization easier.

Concretely, for each support point $\support$, we consider as neighborhood a ball of radius $r$ centered at $\support$, as well as the queries $\querycloud_\support = \{ \query \in \querycloud, \lVert\query-\support\rVert \leq r\}$ falling in this neighborhood. For each such query $\query \in \querycloud_\support$, we create an input to the occupancy decoder which is the concatenation of the latent vector $\latent_\support$ and the local coordinates of the query with respect to the support, i.e., $\query-\support$. The occupancy decoder consists of an MLP and a final sigmoid activation function. For each input vector $\latent_\support \oplus (\query\,{-}\,\support)$, it produces the estimated occupancy at $\query$ given the latent vector at $\support$, denoted $\occupancyest_{\query|\support}$. This predicted occupancy has value in $[0,1]$; it corresponds to an empty (resp.\ full) space if greater (resp.\ less) than 0.5.

\subsection{Reconstruction loss}

Intuitively, we want the loss function to encourage the latent vector $\latent_\support$ of a support point $\support$ to be able to reconstruct everything inside the ball of radius $r$ centered at $\support$. To do so practically, the loss function $\lossrec$  penalizes wrong estimated occupancies for query points $\query$ falling into the ball. Concretely, $\lossrec$ is a binary cross-entropy between the estimated occupancies at the query points $\occupancyest_{\query|\support}$ and the actual sensor-based self-supervised occupancies $\occupancy_\query$:
\begin{equation}
    \lossrec \!=\! \frac{-1}{|\supports|} \! \sum_{\support \in \supports} \! \frac{1}{|\querycloud_\support|} \sum_{\!\!{\query \in \querycloud_\support}\!\!}
    \!\occupancy_\query \log(\occupancyest_{\query|\support}) + (1-\occupancy_\query) \log(1-\occupancyest_{\query|\support})\rlap.
\end{equation}
where $|\supports|$ is the number of support points and $|\querycloud_\support|$ is the number of query points in the ball centered on~$\supports$. Please note that a query point may appear several times in this term, as it may be in the neighborhood of several support points.

\subsection{Particular case of BEV support points}

Current object detectors, such as SECOND~\cite{second} or PV-RCNN~\cite{pvrcnn}, operate a projection on the bird-eye-view (BEV) plane, which has no pre-defined altitude. In this case, instead of considering a ball centered on~$\support$ as query neighborhood $\querycloud_\support$ for support point $\support$, we define $\querycloud_\support$ as the infinite vertical cylinder of radius $r$ centered on~$\support$. As previously, a latent vector $\latent_\support$ should be able to estimate the occupancy of all query points falling in the cylinder.

Please note that the 3D coordinates of $\query$ are still used to compute the input to the occupancy decoder, using a dummy $Z=0$ vertical coordinate to compute the relative location $\query-\support$ that is provided as input to the decoder.

\subsection{Exploiting returned lidar intensity}

Our approach is based on volumetric occupancy estimation given an input point cloud, which is a purely geometric task. However, lidar point clouds often also come with returned lidar intensity at each point, which is widely used to enrich the input features for both semantic segmentation and object detection \cite{cylinder3d,once,pvcnn,pvrcnn,second,pointpillar}. We follow the literature and also input the lidar intensity when available.

Nevertheless, it is not obvious that the network makes the most of both geometry and intensity information. We thus consider here an extra intensity-based loss term, which our ablation study proves beneficial. In this case, we consider a variant of the above presentation where the decoder outputs not only the estimated occupancy~$\occupancyest_{\query|\support}$ but also an estimated intensity~$\intensityest_{\query|\support}$ of the input point $\point$ used to generate query~$\query$.
We then introduce an intensity-recovery loss term  $\lossint$ that penalizes the $\ell_2$ distance between the estimated intensity $\intensityest_{\query|\support}$ of the query point $\query$ and the actual intensity $\intensity_\query = \intensity_\point$ of the corresponding input point $\point$:
\begin{equation}
    \lossint = \frac{1}{|\supports|} \sum_{\support \in \supports} \frac{1}{|\querycloud'_\support|} \sum_{\query \in \querycloud'_\support} ||\intensityest_{\query|\support}-\intensity_\query||_2
\end{equation}
where $\querycloud'_{\support} \subset \querycloud_{\support}$ is the subset of query points $\query \in \querycloud_{\support}$ consisting only of queries $\query_\ifront$ and $\query_\ibehind$ close to sampled points, ignoring queries $\query_\isight$ lying between the sensor and the points, as intensity does not make sense for them. In this setting, the overall loss function is defined as:
\begin{equation}
    \loss = \lossrec + \lambda \lossint
\end{equation}
where $\lambda$ is a weight for balancing the two terms.

%% file: figures/queries.tex
\begin{figure}
    \small
    \centering

    \includegraphics[width=\linewidth]{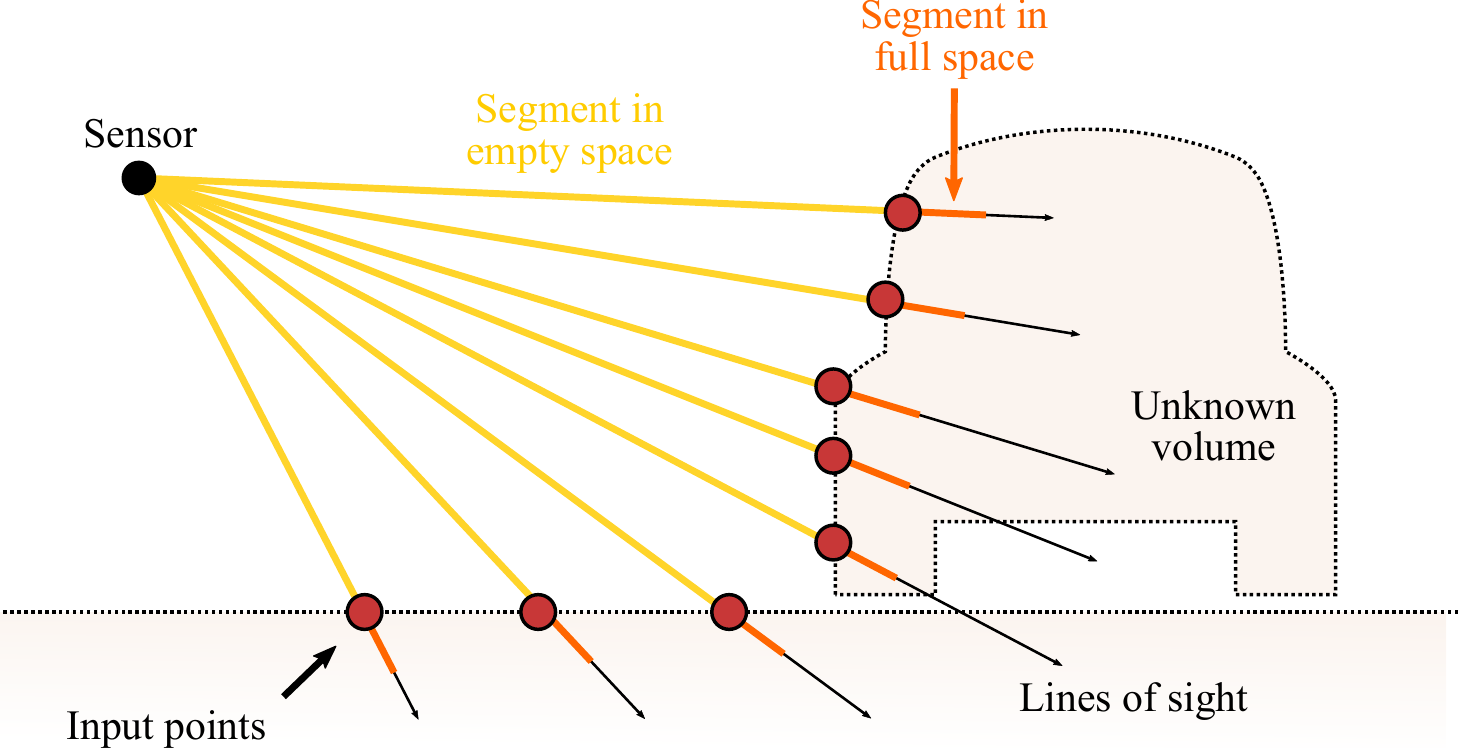}
    (a) Query generation

    \vspace{10pt}
    \includegraphics[width=\linewidth]{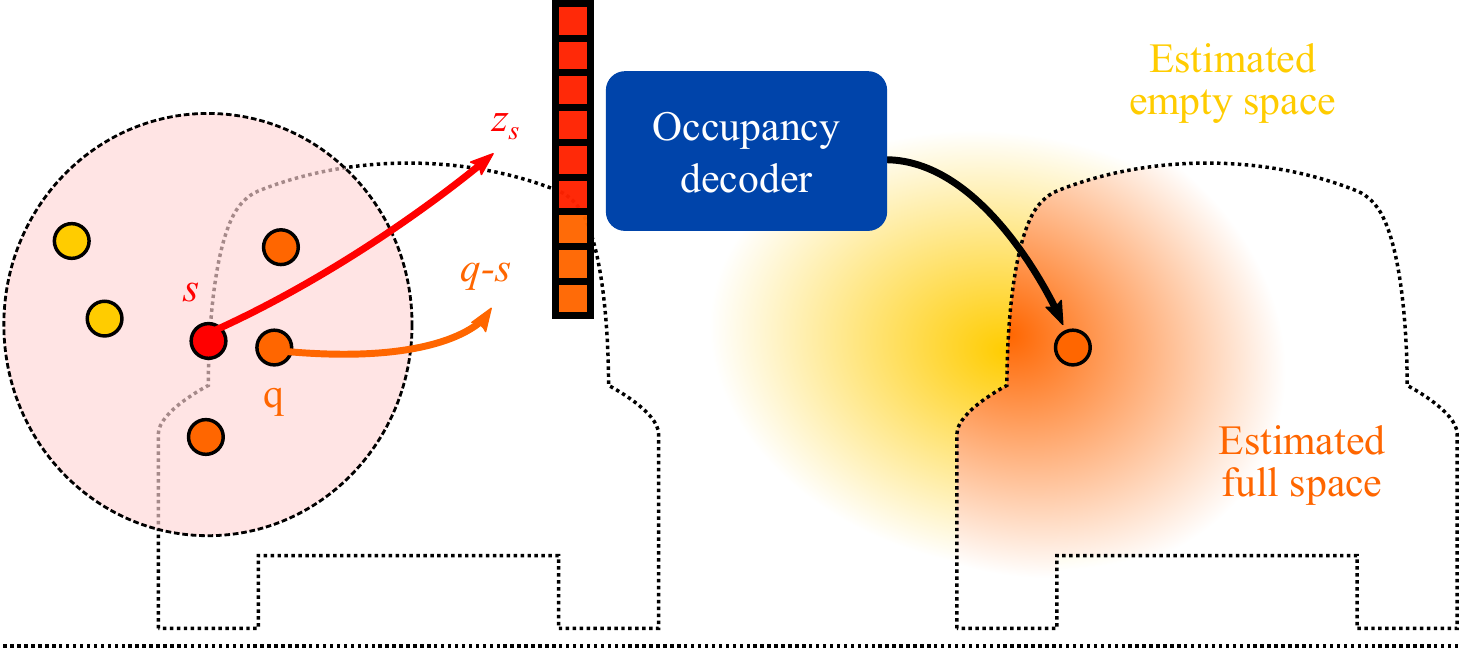}
    (b) Reconstruction process

    \caption{Generation of the queries. The line between the sensor and each points is used to find empty and full points in space. Input features for each query are constructed with the features of support points and the relative coordinates of the query to those points.}
    \label{fig:queries}
\end{figure}

%% file: sections/experiments.tex
\input{figures/qualitative.tex}

\section{Experiments}

To assess our pre-training method, we conduct experiments on both semantic segmentation and object detection.



\subsection{Datasets}

We briefly describe the datasets used for pre-training semantic segmentation (Pre-Seg), downstream semantic segmentation (Seg), pre-training detection (Pre-Det),
\renaud{downstream object detection} (Det). 

\textbf{nuScenes}~\cite{nuscenes} (Pre-Seg, Seg, Pre-Det) is composed of 1000 sequences (train/val/test) acquired with a 32-layer rotative lidar in Boston and Singapore. Points are annotated with 15 classes.
Ablation sets and partial training sets are defined according to~\cite{slidr} (for 0.1\% we use sequence 0392). 

\textbf{SemanticKITTI}~\cite{semantickitti} (Pre-Seg, Seg) contains 22 lidar sequences acquired with a 64-layers Velodyne HDL-64E sensor annotated with 19 labels.
For downstream task, partial training sets are those defined in~\cite{segcontrast}.

\textbf{SemanticPOSS}~\cite{semanticposs} (Seg) is composed of 6 sequences annotated with the same labels as SemanticKITTI.
Contrary to~\cite{segcontrast}, we use the official validation set (sequence 3).

\textbf{LivoxSimu}~\cite{livoxsimu} (Pre-Seg, Seg) is a synthetic dataset simulating 5 Livox Horizon lidars. Points are annotated with 14 labels. The first 90\% of the data is used as training set, the remaining is used as the test set.

\textbf{ONCE} (Pre-Det, Det) contains 1M lidar scenes, most of which are unannotated. Pre-training is done with the small unlabeled set as in \cite{once}, while downstream uses the training and validation splits to train and evaluate the detectors.

\textbf{KITTI\,3D}~\cite{kitti3d} (Pre-Det, Det) is a dataset dedicated to various autonomous driving tasks, including 3D detection. Annotations are provided for $\sim$7.5k frames, in the front camera field of view.
We provide evaluation scores on the moderate difficulty objects with the official 40 recall points $R_{40}$, and 11 recall points $R_{11}$ for comparison purpose.

\textbf{KITTI-360} (Pre-Det) 
is a multiple sensor dataset, including 100k lidar scans in a suburban environment. We use this dataset for detection pre-training purpose.

\subsection{Semantic segmentation}

\textbf{Network architectures.~} 
To evaluate the ability of our approach to generalize to different architectures and for a fair comparison with previous work, we use several backbones.
Experiments are done with: two variants of MinkUNet~\cite{minkowskicnn}, one from~\cite{slidr, pointcontrast}, one from~\cite{segcontrast}, and SPVCNN~\cite{spvcnn}.
The occupancy decoder is a 
4 layer-MLP with interposed ReLU activations and a hidden size of 128, similar to the latent output size.
For segmentation fine-tuning, the occupancy head is removed, as well as the last layer of the backbone, which is replaced by a linear layer with output size corresponding to the number of classes in the dataset.

\textbf{Pre-training.~}
We use the AdamW optimizer with default PyTorch parameters: learning rate $10^{-3}$, $(\beta_1, \beta_2)=(0.9, 0.999)$, $\epsilon=10^{-8}$ and weight decay $0.01$.
For nuScenes and LivoxSimu (resp. SemanticKITTI), we downsample the input to 16k points (resp. 80k), randomly select 2k query points (resp. 4k) per frame and pre-train for 200 epochs with batch size 16 (resp. 50 epochs, with batch size 4).

We set $\delta=10$\,cm. It corresponds more or less to the minimal thickness of objects encountered in outdoor scenes, e.g. poles or human limbs. We use $\lambda=1$ in all our experiments.

\input{tables/table_semseg.tex}

\textbf{Downstream training.~}
Each method's pre-training is evaluated after fine-tuning on downstream tasks.

\textit{Outdoor data.} For fair comparison and in order to setup a simple evaluation protocol, we use the cross-entropy loss and AdamW default PyTorch parameters for all the downstream experiments: learning rate $10^{-3}$, $(\beta_1, \beta_2)=(0.9, 0.999)$, $\epsilon=10^{-8}$ and weight decay $0.01$.
In addition, the learning rate is modulated by a cosine annealing scheduler~\cite{loshchilov2016sgdr} with multiplier ranging from 1 at first epoch to 0.
The fluctuation for any given metric during different runs of the same experiment being significant, we report averages over 5 runs, both for our method, baselines and concurrent works, whenever possible.
For nuScenes and LivoxSimu (resp.\ for SemanticKITTI and SemanticPOSS), we use 16k (resp.\ 80k) input points, batch size 8 (resp.\ 2).
For final score computation, we reproject the labels of the downsampled point cloud on the original point cloud with nearest-neighbor interpolation.
We adapt the number of epochs according to the percentage of training data used.
Note that we use the same number of epochs for all datasets: 1000 for 0.1\%; 500 for 1\%; 100 for 10\%; 50 for 50\% and 30 for 100\%.

\textbf{Ablation studies.~}
Ablation studies are done on nuScenes.
We follow the evaluation procedure of \cite{slidr} where the training set of nuScenes is split in ablation-train and ablation-val sets, and fine-tune using 1\% of the annotations of the ablation-train set.
Doing so ensures that we do not tune method parameters on the validation set which is the set used for comparison to other methods.
We train for 100 epochs.
Ablations are presented in Table~\ref{table:ablations}.

\textit{Context radius.} We study the context radius $r$ to be used for outdoor lidar dataset (Table~\ref{table:ablations}(a)).
$r$ is a crucial parameter.
On the one hand, a small value makes it easier for the network to reconstruct the occupancy, however it will only contain local geometric features without object level shape understanding.
On the other hand, a too large value may cover an area that includes several objects/surfaces, which thus will not favor discrimination between classes/objects in the latent vector.
Intuitively, $r$ should correspond to the scale of the objects that we want to discover in the scene.

\textit{Intensity loss.} Next, we study the impact of intensity both as an input and as a reconstruction objective (Table~\ref{table:ablations}(b)).
It appears that providing intensity to the network already increases the performances (second column).
An additional boost is obtained when using $\lossint$ (third column), enforcing the network to maintain intensity information, which then can easily be used at downstream time.

\textbf{Method evaluation.~}
In this section, we evaluate our method on several datasets and for several sensor types.
Quantitive scores are presented in Table~\ref{table:sem_seg}.

First, we compare \method{} to state-of-the-art methods PointContrast~\cite{pointcontrast}, DepthContrast~\cite{depthcontrast} and SegContrast~\cite{segcontrast} on nuScenes, SemanticKITTI and SemanticPOSS.
Compared to~\cite{segcontrast}, using the default AdamW parameters and training for longer leads to improved performances for all models, including the "no pre-training" setting.
We can observe that performance ranking among the previously cited methods may vary from one dataset to the other, more particularly, DepthContrast benefits a lot from a higher point cloud density.
\method{} outperforms the previous methods for nearly all the configurations (datasets and percentages), e.g., ranking first on nuScenes 1\% by \alex{0.4} \renaud{m}IoU point over PointContrast \renaud{and first} 
on SemanticKITTI 0.1\% by 2.5 points over DepthContrast.
Qualitative examples are presented in Figure~\ref{fig:qualitative_seg}.
We demonstrate here that occupancy reconstruction pre-training method is an efficient and valid alternative to the memory costly contrastive methods.

Second, to hightlight that \method{} is not architecture dependant, we also experiment with SPVCNN~\cite{spvcnn}, a sparse variant of PVCNN~\cite{pvcnn} mixing local point based representation and sparse voxel convolutions.
Our approach presents the same improvement margin over from scratch training than with a MinkUNet.

Third, nuScenes, SemanticKITTI and SemanticPOSS are all datasets created using a rotative lidar.
Even though they were acquired with different sensors, they present similar patterns on the surface, i.e., concentric circles.
We test our approach on the LivoxSimu dataset, where sensors have a very different acquisition patterns.
The MinkUNet network shows a similar behavior as for the rotative sensors, 
\alex{highlighting that our approach can work with different sensors.}

\subsection{Detection}

{\textbf{Network architectures.}}
For detection, we experiment with the commonly used SECOND~\cite{second} and PV-RCNN~\cite{spvcnn} object detectors.
They share the same backbone architecture: a 3D sparse encoder (3D-backbone) made with 3D sparse convolution processing input voxels, and a bird-eye-view (BEV) encoder (2D-backbone) applied after BEV projection.
They mainly differ by the detection heads: SECOND directly applies a region prosal network (RPN) on top of the 2D-backbone, PV-RCNN uses a point-level refinement of the RPN predictions, leading to more accurate boxes and confidence estimation.
We use the OpenPCDet~\cite{openpcdet} implementation of these networks.

\input{tables/table_detection.tex}

\input{tables/table_ablations.tex}

{\textbf{Pre-training.~}}
We pre-train the detection backbone (3D and 2D) using \method{}.
As for semantic segmentation, we train with the default AdamW optimizer parameters, limiting the number of input points to 80k (to prevent high memory peeks), and query points to 4k.
For KITTI3D, we pre-train with batch size 8 for 500 epochs on KITTI3D, 100 on nuScenes and 75 on KITTI-360.
For ONCE, we pre-train for 50 epochs on the $U_{\mathit{small}}$  unannotated set.


{\textbf{Downstream training.~}}
Downstream is done with OpenPCDet~\cite{openpcdet}  for KITTI3D and ONCE~\cite{once} official detection code, in both cases with default settings.

{\textbf{Quantitative evaluation.~}}
Table~\ref{table:detection} displays the scores obtained using our pre-training pipeline.

\textit{Pre-training on the downstream dataset.}
First, we look specifically at methods trained on the target dataset KITTI3D (K) in Table~\ref{table:detection}(a) and on ONCE, Table~\ref{table:detection}(b).
We can observe that we consistently improve over the no-pre-training baseline: +2\% with SECOND, +2.2\% with PV-RCNN on KITTI ($R_{40}$), and by +0.8\% on ONCE.
Compared to literature, we perform on par with Voxel-MAE \cite{voxelmae}.

\textit{Transfering from another dataset.}
Finally, we also pre-trained on KITTI360 and nuScenes to assess the transferability of our pre-training.
We observe first that pre-training on larger datasets, 
leads to higher performances, regardless of the dataset, even if the sensor is not the same.
Then, when considering the choice of the pre-training dataset as a design option, we reach the state of the art on par with ProposalContrast~\cite{proposalcontrast} trained on the Waymo dataset~\cite{waymo}.

%% file: figures/qualitative.tex
\begin{figure*}
    \small
    \centering

    \setlength{\tabcolsep}{1pt}
    \begin{tabular}{cccc|c}
        \includegraphics[width=0.196\linewidth]{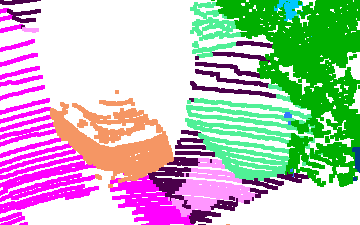} & 
        \includegraphics[width=0.196\linewidth]{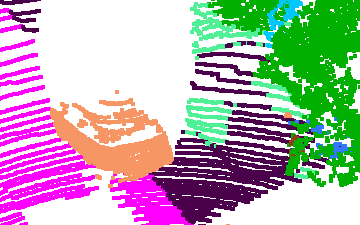} & 
        \includegraphics[width=0.196\linewidth]{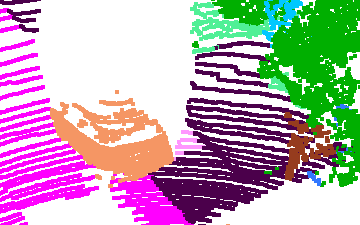} & 
        \includegraphics[width=0.196\linewidth]{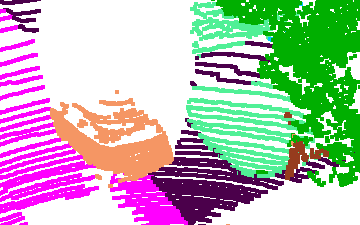} &
        \includegraphics[width=0.196\linewidth]{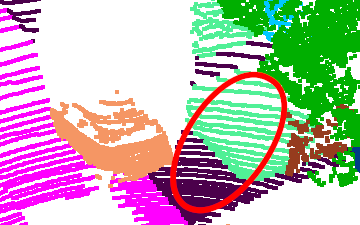} \\
        \includegraphics[width=0.196\linewidth]{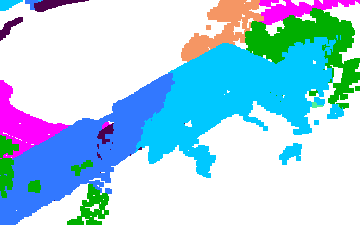} & 
        \includegraphics[width=0.196\linewidth]{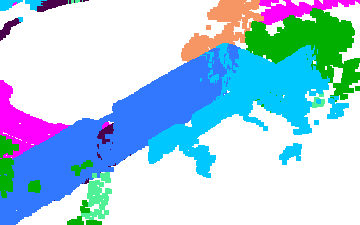} & 
        \includegraphics[width=0.196\linewidth]{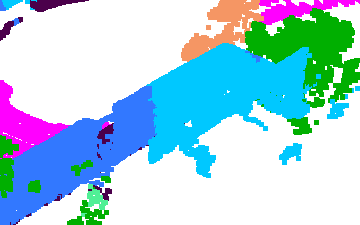} & 
        \includegraphics[width=0.196\linewidth]{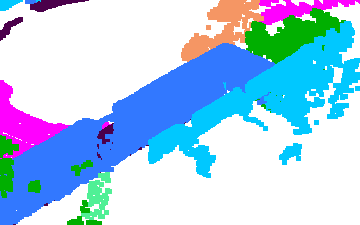} &
        \includegraphics[width=0.196\linewidth]{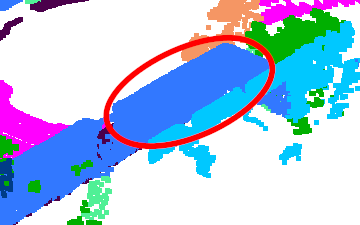} \\
         PointContrast & DepthConstrast & SegContrast & \method{} (ours) & Ground truth \\
    \end{tabular}
\vspace*{-2mm}
    \caption{Qualitative comparison on SemanticKITTI segmentation, fine-tuned with 1\% of training data.}
    \label{fig:qualitative_seg}
\vspace*{-1mm}
\end{figure*}

%% file: tables/table_semseg.tex
\begin{table*}
\small
\setlength{\tabcolsep}{4pt}
\centering

\begin{tabular}{@{}lll|cc|cc|cc|cc|cc@{}}

\toprule
Dataset & Backbone & Method   & \multicolumn{2}{c}{0.1\%} & \multicolumn{2}{c}{1\%} & \multicolumn{2}{c}{10\%} & \multicolumn{2}{c}{50\%} & \multicolumn{2}{c}{100\%}\\

\midrule\midrule
\multirow{6}{*}{nuScenes} &
\multirow{4}{*}{MinkUNet\cite{slidr}}&
No pre-training                     & 21.6 &  & \rthree{35.0} & & 57.3 & & 69.0 & & \rtwo{71.2} & \\
&& PointContrast\cite{pointcontrast} & \rone{27.1} & \tplus{+5.5} & \rtwo{37.0} & \tplus{+2.0} & \rtwo{58.9} & \tplus{+1.6} & \rtwo{69.4} & \tplus{+0.4} & 71.1 & \tminus{-0.1}\\
&& DepthContrast\cite{depthcontrast} & \rthree{21.7} & \tplus{+0.1} & 34.6 & \tminus{-0.4} & \rthree{57.4} & \tplus{+0.1} & \rthree{69.2} & \tplus{+0.2} & \rtwo{71.2} & - \\
 &&\method{} (ours) & \rtwo{26.2} & \tplus{+4.6} & \rone{37.4} & \tplus{+2.4} & \rone{59.0} & \tplus{+1.7} & \rone{69.8} & \tplus{+0.8} & \rone{71.8} & \tplus{+0.6}\\

\arrayrulecolor{black!20}\cmidrule{2-13}\arrayrulecolor{black}
 &\multirow{2}{*}{SPVCNN~\cite{spvcnn}}& 
No pre-training & 22.2  & & 34.4 & & 57.1 & & 69.0 &  & 70.7\\
&&\method{} (ours) & \rone{24.8} & \tplus{+2.6} & \rone{37.4} & \tplus{+3.0} & \rone{58.4} & \tplus{+1.3} & \rone{69.5} & \tplus{+0.5} & \rone{71.3} & \tplus{+0.6}\\



\midrule\midrule
\multirow{7}{*}{SemanticKITTI}&
\multirow{5}{*}{MinkUNet~\cite{segcontrast}}&
No pre-training     & 30.0 &  & 46.2 &  & 57.6 &  & 61.8 &  & 62.7 &  \\
&&PointContrast~\cite{pointcontrast}	& \rthree{32.4} & \tplus{+2.4} & 47.9 & \tplus{+1.7} & \rthree{59.7} & \tplus{+2.1} & \rthree{62.7} & \tplus{+0.9} & \rthree{63.4} & \tplus{+0.7}\\
&&DepthContrast~\cite{depthcontrast}	& \rtwo{32.5} & \tplus{+2.5} & \rtwo{49.0} & \tplus{+2.8} & \rtwo{60.3} & \tplus{+2.7} & \rtwo{62.9} & \tplus{+1.1} & \rone{63.9}	& \tplus{+1.2}\\
&&SegContrast~\cite{segcontrast} & 32.3 & \tplus{+2.3} & \rthree{48.9}  & \tplus{+2.7} & 58.7  & \tplus{+1.1} & 62.1 & \tplus{+0.3} & 62.3 & \tminus{-0.4}   \\
&& \method{} (ours)    & \rone{35.0}  &  \tplus{+5.0} & \rone{50.0} & \tplus{+3.8} & \rone{60.5}  & \tplus{+2.9} & \rone{63.4} & \tplus{+1.6} & \rtwo{63.6} & \tplus{+0.9}\\

\arrayrulecolor{black!20}\cmidrule{2-13}\arrayrulecolor{black}
&\multirow{2}{*}{SPVCNN~\cite{spvcnn}}&
No pre-training & 30.7 &  & 46.6 &  & 58.9 &  & 61.8 &  & 62.7 & \\
&& \method{} (ours) & \rone{35.0} & \tplus{+4.3} & \rone{49.1} & \tplus{+2.5} & \rone{60.6} & \tplus{+1.7} & \rone{63.6} & \tplus{+1.8} & \rone{63.8} & \tplus{+1.1}\\

\midrule\midrule
&\multirow{5}{*}{MinkUNet~\cite{segcontrast}}& 
No pre-training    & 36.9 &  & 46.4 &  & 54.5 &  & 55.3 &  & 55.1 &  \\
SemanticPOSS && PointContrast~\cite{pointcontrast} & 39.3 & \tplus{2.4} & 48.1 & \tplus{1.7} & 55.1 
& \tplus{+0.6} & \rthree{56.2} & \tplus{+0.9} & 56.2 & \tplus{+1.1}\\
(pre-training&& DepthContrast~\cite{depthcontrast} & \rthree{39.7} & \tplus{+2.8} & \rthree{48.5} & \tplus{+2.1} & \rtwo{55.8} & \tplus{+1.3} & 56.0 & \tplus{+0.7} & \rtwo{56.5} & \tplus{+1.4}\\
SemanticKITTI)&& SegContrast~\cite{segcontrast} & \rone{41.7} & \tplus{+4.8} & \rtwo{49.4} & \tplus{+3.0} & \rthree{55.4} & \tplus{+0.9} & \rtwo{56.2} & \tplus{+0.9} & \rthree{56.4} & \tplus{+1.3}\\
&& 
\method{} (ours)    & \rtwo{40.7} & \tplus{+3.8} & \rone{49.6} & \tplus{+3.2} & \rone{55.8} & \tplus{+1.3} & \rone{56.4} & \tplus{+1.1} & \rone{56.7} & \tplus{+1.6}\\

\midrule\midrule
\multirow{2}{*}{LivoxSimu}&
\multirow{2}{*}{MinkUNet~\cite{slidr} }& 
No pre-training & 48.0 &  & 63.8 &  & 66.7 &  & 68.5 &  & 68.9 & \\
&&\method{} (ours) & \rone{52.6} & \tplus{+4.6} & \rone{65.5} & \tplus{+1.7} & \rone{67.8} & \tplus{+1.1} & \rone{69.6} & \tplus{+1.1} & \rone{69.7} & \tplus{+0.8}\\

\bottomrule

\multicolumn{13}{l}{\textit{Results averaged over 5 runs. \renaud{Individual} run details and standard deviation in the supplementary material.}}\\
\end{tabular}

\caption{Semantic segmentation results. We report the mIoU \renaud{(\%)} of fine-tuned models on four different datasets, while varying the amount of annotated data, the pre-training dataset and the architecture. We compare \method{} against a non-pre-trained baseline and concurrent work.}
\label{table:sem_seg}
    
\end{table*}

%% file: tables/table_detection.tex
\begin{table}[!ht]
\small
\setlength{\tabcolsep}{4pt}

\centering

\begin{tabular}{@{}l|c|ccc|cc@{}}
\multicolumn{7}{@{}l}{(a) KITTI3D~\cite{kitti3d}, validation set, moderate difficulty.} \\
\toprule
 Method & Data. & Cars & Ped. & Cycl. & mAP & Diff.\\

\midrule
\multicolumn{7}{@{}l}{SECOND - $R_{40}$ \renaud{metric}}\\
 No pre-training& -  & 81.50 & 48.82 & 65.72 & 65.35 &              \\
 \greyrule{\midrule}
                            & K  & \rone{81.97} & 51.93 & \rtwo{69.14} & 67.68 & \tplus{+2.33}\\
                            &K360& \rtwo{81.79} & \rtwo{52.45} & \rone{70.68} & \rone{68.31} & \tplus{+2.96} \\
\multirow{-4}{*}{\method{} (ours)} & NS & \rtwo{81.78} & \rone{54.24} & 68.19 & \rtwo{68.07} & \tplus{+2.72} \\

\midrule
\multicolumn{7}{@{}l}{SECOND - $R_{11}$ \renaud{metric}}\\

No pre-training             & -  & 78.62 & 52.98 & 67.15 & 66.25 & \\
Voxel-MAE~\cite{voxelmae}   & K  & \rone{78.90} & 53.14 & 68.08 & 66.71 & \tplus{+0.46} \\
\greyrule{\midrule}
                            & K  & 78.78 & 53.57 & \rtwo{68.22} & 66.86 & \tplus{+0.61}\\ 
                            &K360& 78.63 & \rtwo{54.23} & \rone{69.35} & \rone{67.40} & \tplus{+1.15}\\
\multirow{-4}{*}{\method{} (ours)} & NS & \rtwo{78.65} & \rone{55.17} & 68.05 & \rtwo{67.29} & \tplus{+1.04}\\

\midrule
\multicolumn{7}{@{}l}{PV-RCNN - $R_{40}$ \renaud{metric}}\\

No pre-training                     & -  & 84.50 & 57.06 & 70.14 & 70.57 &  \\
 STRL~\cite{strl}                   & K  & 84.70 & 57.80 & 71.88 & 71.46 & \tplus{+0.89} \\
GCC-3D~\cite{gcc3d}                 & NS & -     & -     & -     & 70.75 & \tplus{+0.18} \\
GCC-3D~\cite{gcc3d}                 & W  & -     & -     & -     & 71.26 & \tplus{+0.69} \\
PointCont.~\cite{pointcontrast} 
                                    & W  & 84.18 & 57.74 & 72.72 & 71.55 & \tplus{+0.98} \\
Prop.Cont.~\cite{proposalcontrast}   
                                    & W  & \rtwo{84.72} & \rone{60.36} & 73.69 & \rtwo{72.92} & \tplus{+2.35} \\
\greyrule{\midrule}
                                    & K  & \rtwo{84.72} & 58.49 & \rone{75.06} & 72.76 & \tplus{+2.19}\\
                                    &K360& 84.68 & \rtwo{60.16} & 74.04 & \rone{72.96} & \tplus{+2.39} \\
\multirow{-4}{*}{\method{} (ours)}  & NS & \rone{84.86} & 57.76 & \rtwo{74.98} & 72.53 & \tplus{+1.98} \\

\midrule
\multicolumn{7}{@{}l}{PV-RCNN - $R_{11}$ \renaud{metric}}\\

No pre-training             & -  & 83.61 & 57.90 & 70.47 & 70.66 & \\
Voxel-MAE~\cite{voxelmae}   & K  & \rone{83.82} & \rtwo{59.37} & 71.99 & 71.73 & \tplus{+1.07} \\
\greyrule{\midrule}
                            & K  & 83.67 & 58.48 & 73.74 & 71.96 & \tplus{+1.30}\\
                            &K360& 83.39 & \rone{60.83} & \rtwo{73.85} & \rone{72.69} & \tplus{+2.03}\\
\multirow{-4}{*}{\method{} (ours)} & NS & \rtwo{83.77} & 58.49 & \rone{74.35} & \rtwo{72.20} & \tplus{+1.54}\\
\bottomrule
\multicolumn{1}{l}{~}\\[-6pt]
\multicolumn{7}{@{}l}{(b) ONCE~\cite{once}, validation set, SECOND detector, \renaud{ONCE metric}.}\\
\toprule
 Method & Data. & Cars & Ped. & Cycl. & mAP & Diff.\\
\midrule
No pre-training                     & - & 71.19 & 26.44 & 58.04 & 51.89 &  \\
BYOL~\cite{byol}                    & O$_\text{s}$ & 68.02 & 19.50 & 50.61 & 46.04 & \tminus{-5.85} \\
PointCont.~\cite{pointcontrast}  & O$_\text{s}$ & 71.07 & 22.52 & 56.36 & 49.98 & \tminus{-1.91} \\
SwAV~\cite{swav}                    & O$_\text{s}$ & \rtwo{72.71} & \rtwo{25.13} & 58.05 & 51.96 & \tplus{+0.07} \\
DeepCluster~\cite{deepcluster}      & O$_\text{s}$ & \rone{73.19} & 24.00 & \rone{58.99} & \rtwo{52.06} & \tplus{+0.17} \\
\greyrule{\midrule}
\method{} (ours)                    & O$_\text{s}$ & 71.73 & \rone{28.16} & \rtwo{58.13} & \rone{52.68} & \tplus{+0.79} \\
\bottomrule
\multicolumn{7}{@{}l@{}}{\textit{\qquad \renaud{Datasets:} KITTI3D (K), KITTI-360 (K360), nuScenes (NS),}}\\
\multicolumn{7}{@{}l@{}}{\textit{\qquad ONCE Small (O$_\text{s}$), Waymo (W).}}\\
\end{tabular}
\vspace*{-2mm}
\caption{Detection results on KITTI3D (a) and ONCE (b). We report \renaud{AP (\%)} 
and the dataset used for pre-training each method.}
\label{table:detection}
\vspace*{-1mm}
\end{table}

%% file: tables/table_ablations.tex
\begin{table}
    \small
    \centering
    (a) Reconstruction radius (with intensity and $\lossint$)
    \begin{tabular}{c|cccc}
        \toprule
        Radius (m) & 0.5 & 1 & 2 & 4 \\
        \midrule
        mIoU \renaud{(\%)} & 37.6 & \textbf{38.4} & 38.2 & 36.4\\
        \bottomrule
    \end{tabular}
    
    \vspace*{3mm}

    (b) \mbox{Intensity for pre-training (with radius=1.0\,m)} 
    \begin{tabular}{c|ccc}
        \toprule
        Input intensity    &  \xmark & \cmark & \cmark\\
        Loss $\lossint$    &  \xmark & \xmark & \cmark\\
        \midrule
        mIoU \renaud{(\%)}               &   36.4  & 38.2   & \textbf{38.4} \\
        \bottomrule
    \end{tabular}
    
    \caption{\renaud{Parameter study for radius $r$~(a) and ablation study of intensity usage~(b)}, 
    as input and objective \renaud{for pre-training}. Evaluation is on the ablation-val set of nuScenes, training 100 epochs.}    
    \label{table:ablations}
    \vspace*{-1mm}
\end{table}

%% file: sections/conclusion.tex
\section{Conclusion}

In this work we investigate the use of occupancy reconstruction as a pretext task for self-supervision on point cloud. We show that a supervision signal for occupancy estimation can directly be inferred from the sensor information.
The resulting method is conceptually simple and can be trained with limited resources (single 16GB memory GPU).
\method{} can be used for semantic segmentation as well as for detection, and provides clear benefits on tested architectures and datasets.
It outperforms contrastive methods on semantic segmentation and is able to perform on par with state-of-the-art detection methods specifically designed for the task.

In the footsteps of input reconstruction approaches, we show that geometric tasks, here estimating the occupancy, are meaningful alternatives to contrastive learning and masked autoencoders. \renaud{Future work include studying combinations with contrast-based and completion-based approaches.}


\noindent\paragraph{Acknowledgements:}
\alex{
This work was supported in part by the French Agence Nationale de la Recherche (ANR) grant MultiTrans (ANR21-CE23-0032).
This work was performed using HPC resources from GENCI–IDRIS (Grants 2021-AD011012883 and 2022-AD011012883R1). 
}

%% file: sections/appendix.tex
\setcounter{figure}{0}
\setcounter{table}{0}
\renewcommand{\thefigure}{A\arabic{figure}}
\renewcommand{\thetable}{A\arabic{table}}

\doparttoc 
\faketableofcontents 
\addcontentsline{toc}{section}{Appendix} 
\part{Appendix} 
\hypersetup{linkcolor=black}
\parttoc 




\section{Experimental settings}

In this section, we give additional information for reproducibility purpose.
%
{Additionally, the code is publicly available at \href{https://github.com/valeoai/ALSO}{github.com/valeoai/ALSO} under an open source license.}

\subsection{Self-supervision}

\subsubsection{Occupancy decoder}

The occupancy decoder, presented in Figure~\ref{fig:supp_decoder_arch}, is a four-layer MLP.
It takes as input the concatenation of the latent vector and the local coordinates of the query point $\query$ with respect to the support point $\support$.
We use ReLU activations, and the hidden size of the MLP is set to 128, which is the size of the latent space. 

We also provide a code sample for the decoder in Listing~\ref{code:decoder}.
The code is written in Python, using PyTorch \cite{Pytorch2019NIPS} and PyTorch Geometric \cite{pytorchgeometric} for neighborhood computation.

\subsubsection{Data transformation.}

We detail here the transformations of the data used at pre-training.

\input{figures/supp_decoder.tex}

\input{tables/code_decoder.tex}

\paragraph{Semantic segmentation pre-training.}
The voxel-size used in the sparse convolution backbone is set to 0.1\,m for nuScenes/LivoxSimu and 0.05\,m for SemanticKITTI/SemanticPOSS.

\paragraph{Detection pre-training.}
The voxel sizes are those used in OpenPCDet \cite{openpcdet} and ONCE \cite{once} for the different backbones.
On KITTI, the voxel size is 0.05\,m on the horizontal plane and 0.1\,m in the vertical direction.
On ONCE, the voxel size is 0.1\,m on the horizontal plane and 0.2\,m in the vertical direction.

\paragraph{Data augmentations.}
We apply classical point cloud data augmentations: random rotation around the z-axis as well as random flipping of the other axes.

\paragraph{Hardware configuration.}
For all our pre-traininings, we use a single Nvidia V100 16GB GPU.

\subsection{Details for finetuning on downstream tasks}

\paragraph{Weight initialization.}

For semantic segmentation, we initialize all the backbone's weights with the pre-trained weights, except for the last layer (used for point-wise classification) which is randomly initialized.
Then we finetune all the layers with the same learning rate, as described in the main paper.

For detection, we initialize the network's weights using the pre-trained weights, both for the sparse backbone and the dense BEV layers. 
These backbone and dense layers are finetuned along with the SECOND/PV-RCNN detectors on the task of supervised 3D object detection.

\paragraph{Hardware configuration.}

Semantic segmentation and KITTI detection downstream experiments are done on a single Nvidia RTX2080Ti 11GB GPU.
For ONCE detection, we used 8 Nvidia V100 16GB GPUs.

\section{Analysis of the latent space}

\input{figures/supp_latent.tex}

In the main paper, we showed that using a self-supervised geometric pretext task is powerful to pre-train a backbone for both semantic segmentation and object detection.
In this section, we look at the structure of the underlying latent space, learned using \method{}.

\subsection{Natural clusters in latent space}


We randomly select 15 nuScenes point clouds from which we select at most 2000 points of each semantic class. We gather the self-supervised latent vectors corresponding to these points and embed them in a 2-dimensional space using t-SNE \cite{van2008visualizing}. 

Figure~\ref{fig:supp_latent} presents the result of this t-SNE embedding, where the colors encode the semantic classes.
We notice that points belonging to the same class tends to be clustered together in this representation. When restricting the analysis to the car class and selecting neighbors in this representation, we notice that the corresponding cars are captured from the same point-of-view (rear right from the ego vehicle), indicating that the latent space probably tends to group together objects sharing the same apparent geometry.  

\subsection{Expected geometric properties visible in the latent space.}

\paragraph{Surface orientation.}
In Figure~\ref{fig:supp_latent2}\,(a), we differentiate points belonging to horizontal flat surfaces (driveable surface, sidewalk, terrain and other flat surfaces) as opposed to points on objects usually sampled from the side, i.e., where surface is vertical (buildings, pedestrian...). We remark we can almost linearly separate these two sets of points in the t-SNE representation of the latent space, showing that the network rely on low-level geometric features, e.g., rough normal estimation, to solve the pretext task.

\input{figures/supp_latent2.tex}

\paragraph{Symmetric occupancy map.}

In Figure~\ref{fig:supp_latent2}\,(b), we differentiate points with positive and negative $x$-coordinate, i.e., point on the right and on the left of the ego-vehicle. Again, we notice that these two sets of points are quite well separated in the t-SNE representation. 
This is explained by the fact the occupancy reconstruction task produces oriented surfaces. Two similar objects but located on different side of the road exhibit symmetric occupancy maps w.r.t. the $(y,z)$ plane (see Figure~\ref{fig:appendix_symm}), yielding different representations in the latent space.

\input{figures/supp_symmetry.tex}

\section{Additional \renaud{parameter and alternative} studies}

\paragraph{$\delta$ parameter.}
\alex{In our approach, a location at a
random distance in $[0, \delta]$ behind an observed point is deemed
occupied. We argue this heuristics, also successfully used in~\cite{Raphael}, is simple and correct often enough although, as any heuristics, it can occasionally be wrong, just as \renaud{random} negatives in contrastive learning are also sometimes wrong.
Importantly, our heuristics is stable across a significant range of values for $\delta$, as shown in Table~\ref{table:ablations_supp}\,(a). In our experiments, we uniformly chose $\delta = 0.1$\,m as a kind of minimal thickness of the sort of objects we want to perceive. But, as visible
in this table, there may be slightly more beneficial values depending on the dataset, e.g., $0.2$\,m for nuScenes.
}

\paragraph{Decoder head.}
\alex{
Local information is good for accurate surface reconstruction~\cite{poco,convonet}: it allows each point feature to focus on local geometric details, as the decoder aggregates \renaud{contextual information to predict occupancy}. 
Instead, in ALSO, we force each individual point to know how to reconstruct its entire neighborhood on its own. Doing so yields \renaud{single} point features that are more context-aware, hence
with a more semantic flavor, at the \renaud{possible} cost of a \renaud{less-accurate} 
surface reconstruction. In Table~\ref{table:ablations_supp}\,(b), we provide
\renaud{an additional study to compare different reconstruction heads:} 
POCO head~\cite{poco}, ball search + average/maximum pooling. It shows that limiting the expressiveness of the decoder to a per-point MLP \renaud{(\method{} head)} helps self-supervision.
}

\input{tables/table_ablations_supp.tex}

\section{Additional visualizations}

We also provide additional visualizations on Figure~\ref{fig:supp_qualitative_occ_seg}. They are produced similarly to Figure~1 in the main paper.
To produce these aggregated views, we compute the occupancy in a 1-meter radius \renaud{ball} from the input points by randomly picking query points in this ball.
Each inside point is then labeled with the estimated class of the closest input point.
These estimated classes are obtained with the downstream model finetuned for semantic segmentation.
Visualization are provided for SemanticKITTI, nuScenes and LivoxSimu.

\input{figures/supp_occupancy.tex}

\section{Semantic segmentation: experiment scores details}

All the scores for the ablation study and the semantic segmentation experiments are averaged over 5 runs in the main paper. We provide here all the individual score:
in Table~\ref{tab:appendix_ablations} for the ablation study,
in Table~\ref{tab:appendix_nuscenes} for the experiments on nuScenes,
in Table~\ref{tab:appendix_semantickitti} for the experiments on SemanticKITTI,
in Table~\ref{tab:appendix_semanticposs} for the experiments on SemanticPOSS and in Table~\ref{tab:appendix_livox} for the experiments on LivoxSimu. In each table, we highlight in bold the best run, and report the average score (same as in the paper) as well as the the standard deviation.

\input{tables/table_supp_ablations.tex}

\input{tables/table_supp_nuscenes.tex}

\input{tables/table_supp_semantickitti.tex}

\input{tables/table_supp_semanticposs.tex}

\input{tables/table_supp_livox.tex}

\section{3D detection: experiment scores details}

Last, we provide more detailed scores for the experiments on 3D object detection.

On KITTI-3D detection, we provide in Table~\ref{tab:appendix_detection_kitti3d} scores for the different official metrics (2D object detection, bird's-eye-view detection, 3D object detection, orientation similarity).
We also report the scores for the easy and hard categories (in the main paper, the reported scores correspond to the moderate difficulty category).
Cells are colored according to the gain obtained using self-supervised weight initialization.

\method{} offers a performance boost on pedestrian and cyclists no matter what the metric is. However, on the car class, the gain is reduced, probably because the car class is already performing well.

\input{tables/table_supp_kitti3d.tex}

As in~\cite{once}, we report in Table~\ref{table:supp_once} the per-distance performance for each of the three classes of interest.
Our approach performs on par with the proposed baselines, including DeepCluster \cite{deepcluster}.
The performance boost is mainly due to the good detection performance of pedestrians between 0 and 50 meters.

\input{tables/table_supp_once.tex}

\addtocontents{toc}{%
  \protect\endgroup%
}

%% file: figures/supp_decoder.tex
\begin{figure}
    \centering
    \includegraphics[width=\linewidth]{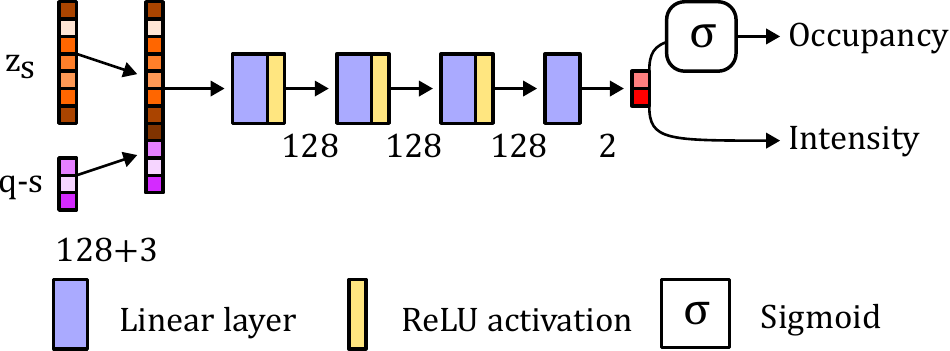}
    \caption{Decoder architecture.}
    \label{fig:supp_decoder_arch}
\end{figure}

%% file: tables/code_decoder.tex
\begin{table}[!ht]
    
\begin{lstlisting}[language=Python, caption={Pseudo-code of the decoder code using PyTorch syntax.}, label=code:decoder]
import torch
from torch.nn as Linear, ReLU
from torch.nn.functional import binary_cross_entropy_with_logits as bce_loss, l1_loss
from functools import partial
from torch_geometric.nn import radius

# l_size: latent_size
# o_size: output size
# r: neighborhood radius

class OccupancyDecoder(torch.nn.Module):

  def __init__(self, l_size=128, o_size=2, r=1):
    super().__init__()

    mlp = []
    for i in range(3):
        mlp.append(Linear(l_size, l_size))
    mlp.append(Linear(l_size, o_size))
    self.mlp = torch.nn.Sequential(*mlp)
    self.ball_search = partial(radius, r=r)
    self.r = r
    
  def forward(self, data):

    # get data from input dictionary
    pos_support = data["pos_support"]
    batch_support = data["batch_support"]
    pos_query = data["pos_query"]
    batch_query = data["batch_query"]
    latent = data["latent"]

    ## NEIGHBORHOOD SEARCH - LOCAL COORDINATES
    row, col = self.ball_search(x=pos_support, 
        y=pos_query, batch_x=batch_support,
        batch_y=batch_query)
    pos_local = pos_query[row] - pos_support[col]
    l_local = latent[col]

    ## OCCUPANCY ESTIMATION
    x = torch.cat([l_local, pos_local], dim=1)
    x = self.mlp(x)
    occ_preds, i_preds = x[:, 0], x[:, 1]

    ## LOSS COMPUTATION
    occ_gt = data["query_occupancy"][row]
    occ_loss = bce_loss(occ_preds, occ_gt)

    i_gt = data["query_intensity"][row]
    i_mask = (i_gt >= 0)
    i_loss = l1_loss(i_preds[i_mask], 
                     i_gt[i_mask])
        
    return occ_loss + i_loss
\end{lstlisting}
\end{table}

%% file: figures/supp_latent.tex
\begin{figure}
    \centering
    \includegraphics[width=\linewidth]{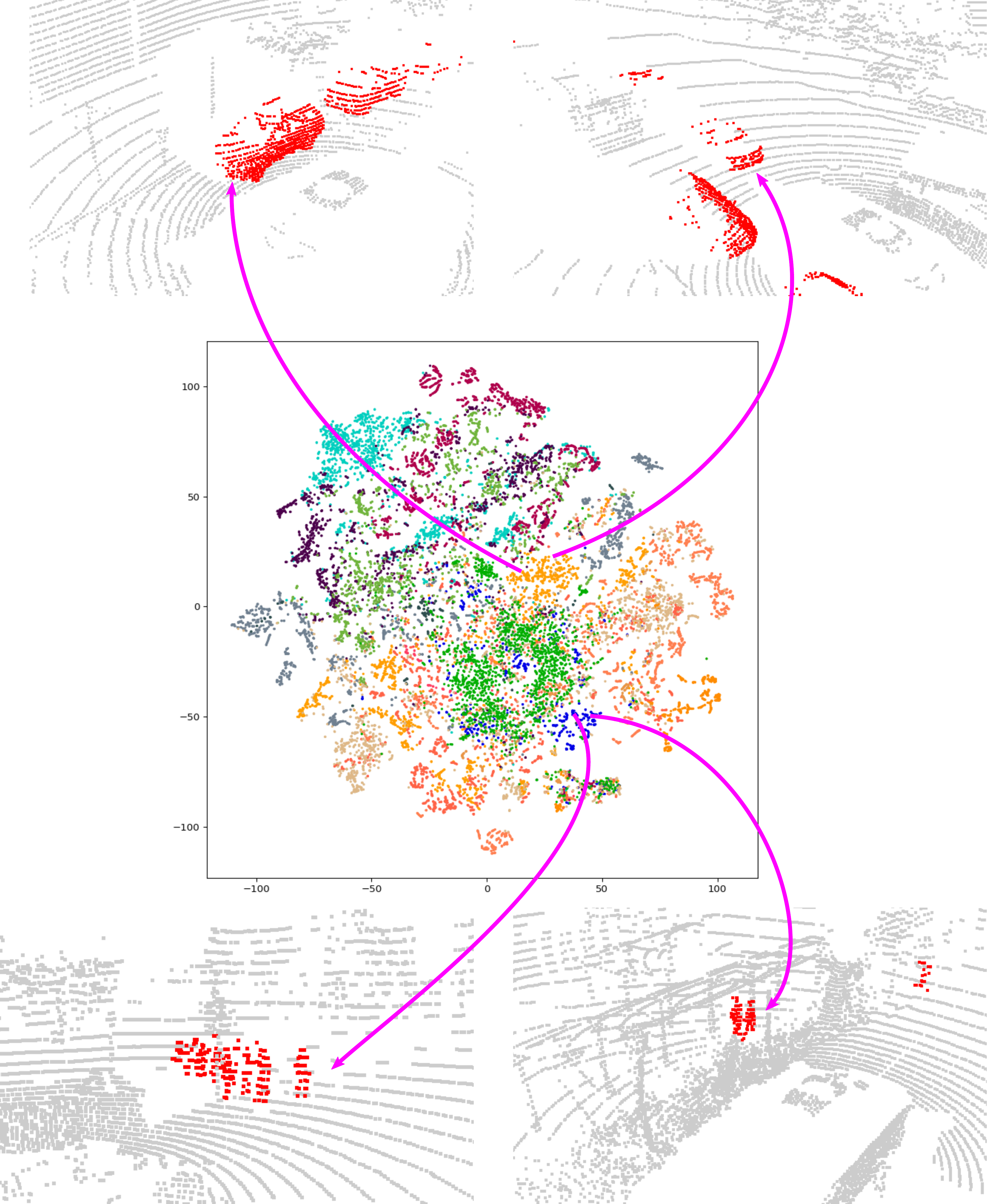}
    \caption{Latent space visualization.}
    \label{fig:supp_latent}
\end{figure}

%% file: figures/supp_latent2.tex
\begin{figure}[!ht]
    \small
    \centering
    \setlength{\tabcolsep}{1pt}
    \begin{tabular}{@{}cc@{}}
    \includegraphics[width=0.49\linewidth]{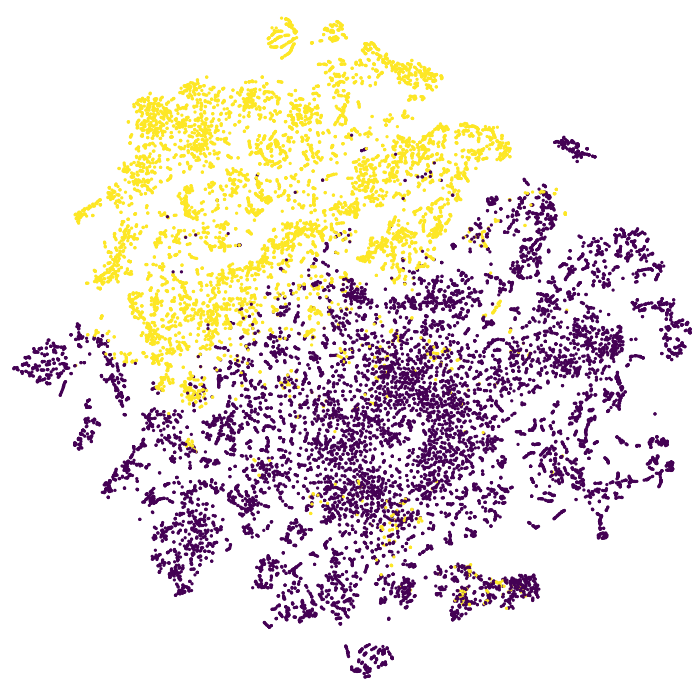} & 
    \includegraphics[width=0.49\linewidth]{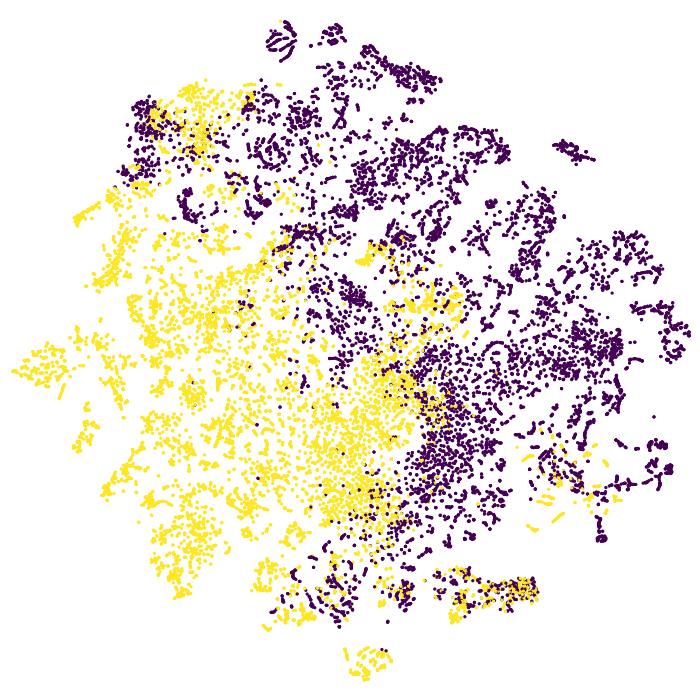} \\
    (a) Flat \renaud{horizontal} surfaces (yellow) & (b) Side w.r.t. ego \renaud{orientation:}
    \\
    vs vertical objects (purple) & right (yellow) and left (purple)\\
    \end{tabular}
    \caption{Geometric structure of the latent space (nuScenes).}
    \label{fig:supp_latent2}
\end{figure}

%% file: figures/supp_symmetry.tex
\begin{figure}
    \centering
    \includegraphics[width=0.8\linewidth]{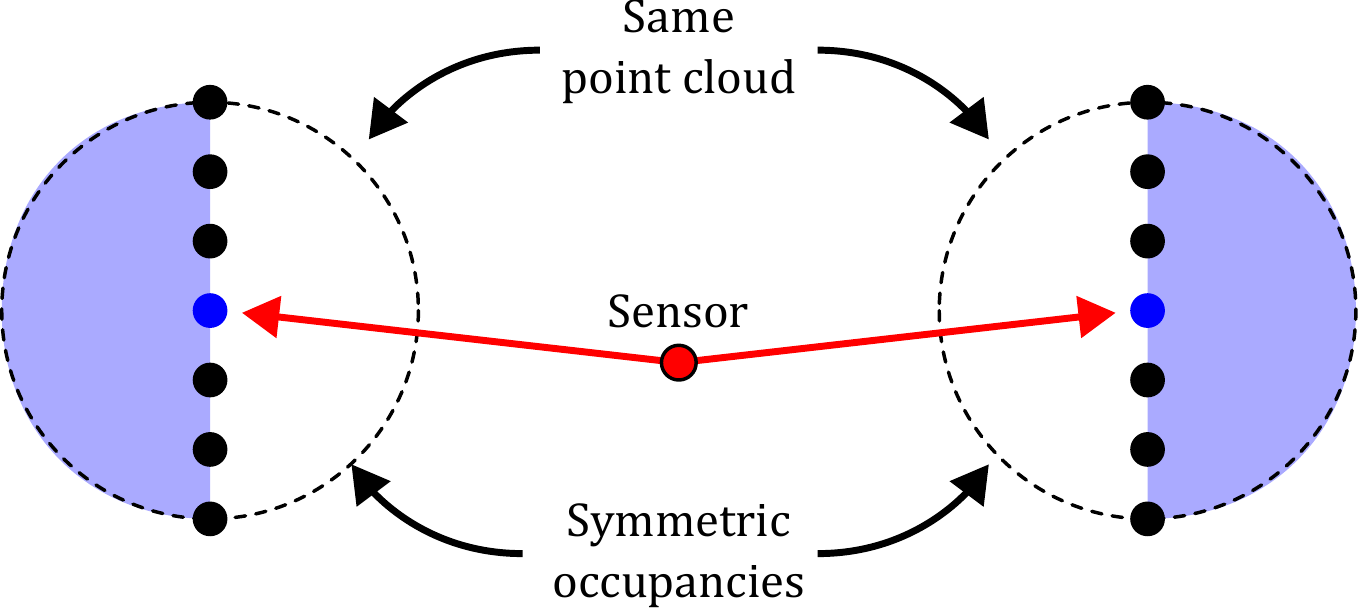}
    \caption{Symmetry of occupancy.}
    \label{fig:appendix_symm}
\end{figure}

%% file: tables/table_ablations_supp.tex
\begin{table}
    \small
    \centering
    \setlength{\tabcolsep}{3pt}

    \begin{tabular}{c|ccccc}
    \multicolumn{6}{l}{(a) $\delta$ parameter study}\\
    \toprule
    $\delta$ (m)   &  0.05     & 0.1   & 0.2       & 0.4       & 0.8\\
    \midrule
     mIoU           & 38.3   & 38.4  & 38.7   & 38.7   & 38.1\\ 
     \bottomrule
    \end{tabular}

    \vspace{2mm}
    
    \begin{tabular}{c|c|c|c|c}
    \multicolumn{5}{l}{(b) Decoder head \renaud{alternatives}}\\
    \toprule
     Decoder &  POCO & Ball + & Ball + & ALSO \\
    head & (Knn + Att.)  & Avg. & Max. & Ball per point \\
    \midrule
    mIoU\%  &  33.8    & 35.7 & 35.8 & 38.4 \\
    \bottomrule
    \end{tabular}
        
    \caption{\renaud{Parameter} study for $\delta$~(a) and \renaud{alternative study} for the decoder head~(b) during the pre-training. Experiments are evaluated on the ablation-val set of nuScenes.}
    \label{table:ablations_supp}
\end{table}

%% file: figures/supp_occupancy.tex
\begin{figure*}
    \small
    \centering
    \setlength{\tabcolsep}{1pt}

    \begin{tabular}{@{}cc|cc@{}}
        \includegraphics[width=0.245\linewidth]{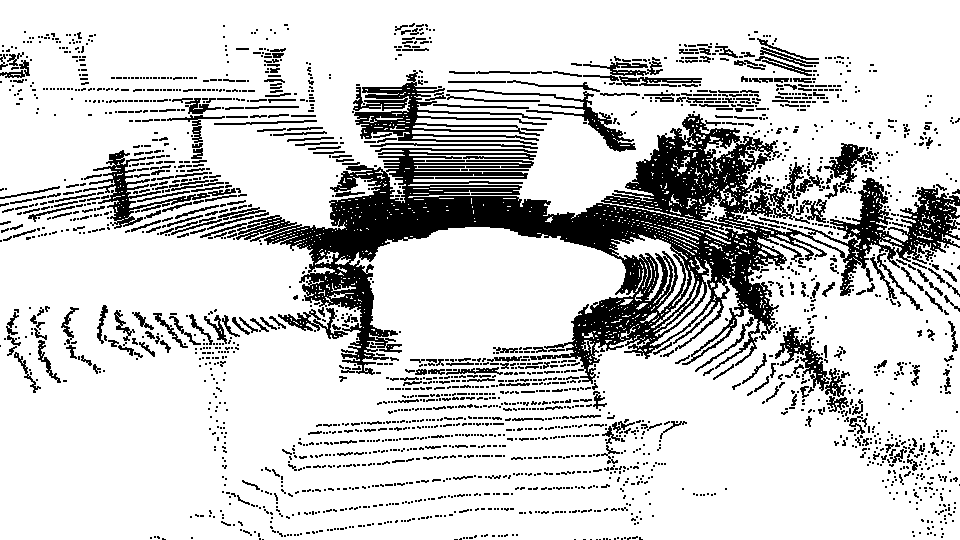} &
        \multicolumn{1}{c}{\includegraphics[width=0.245\linewidth]{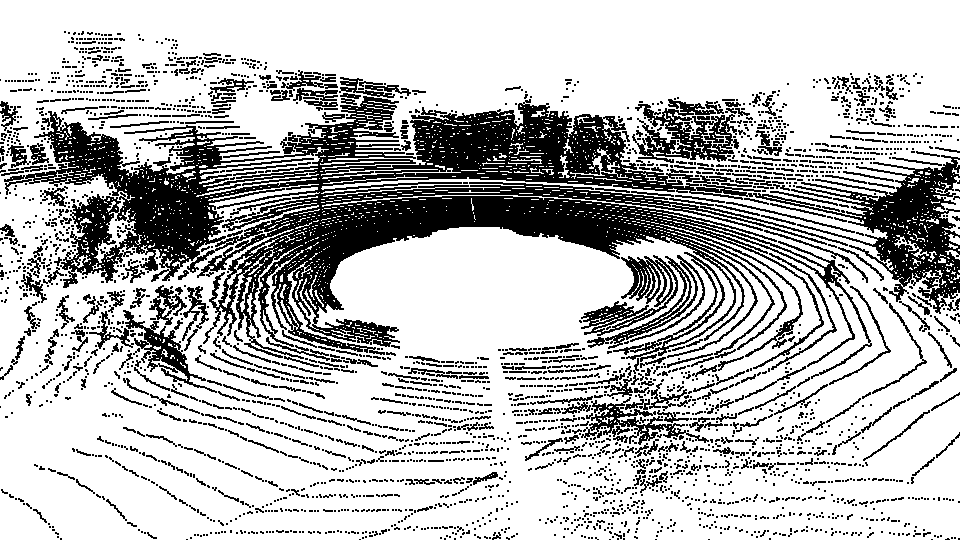}} &
        \includegraphics[width=0.245\linewidth]{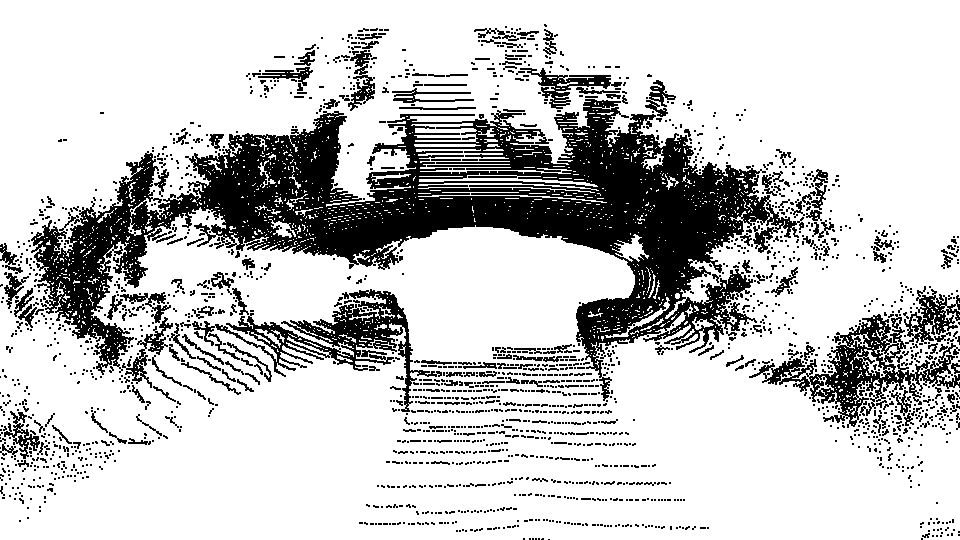} & 
        \includegraphics[width=0.245\linewidth]{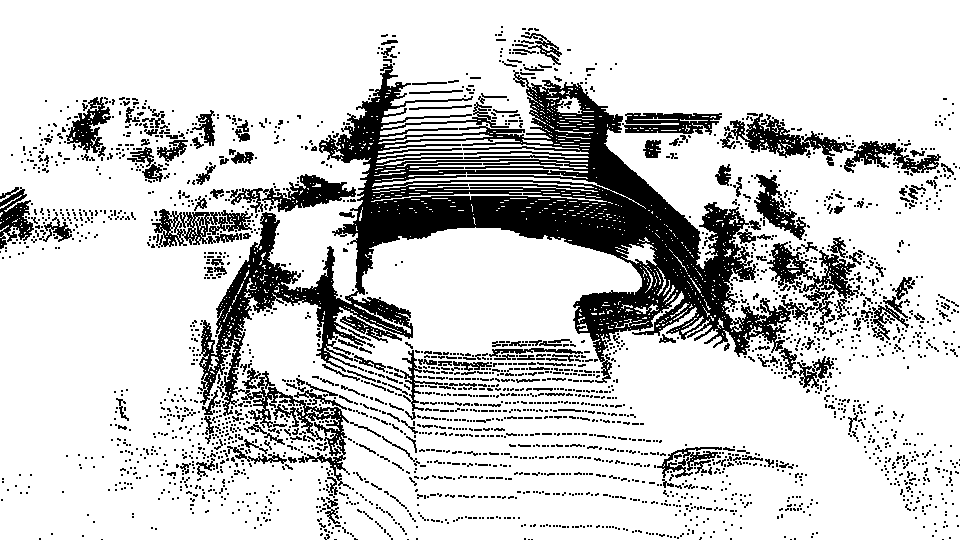} \\
        
        \includegraphics[width=0.245\linewidth]{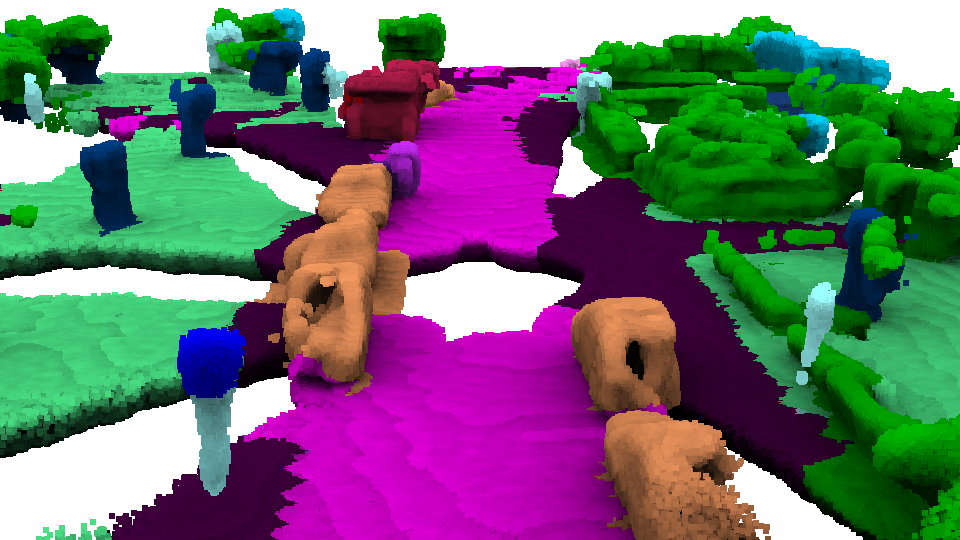} &
        \multicolumn{1}{c}{\includegraphics[width=0.245\linewidth]{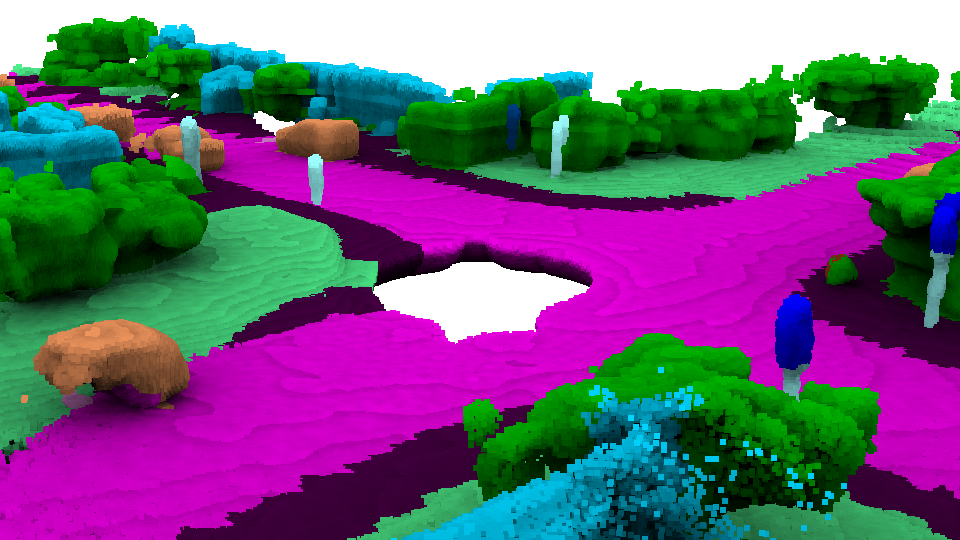}} & 
        \includegraphics[width=0.245\linewidth]{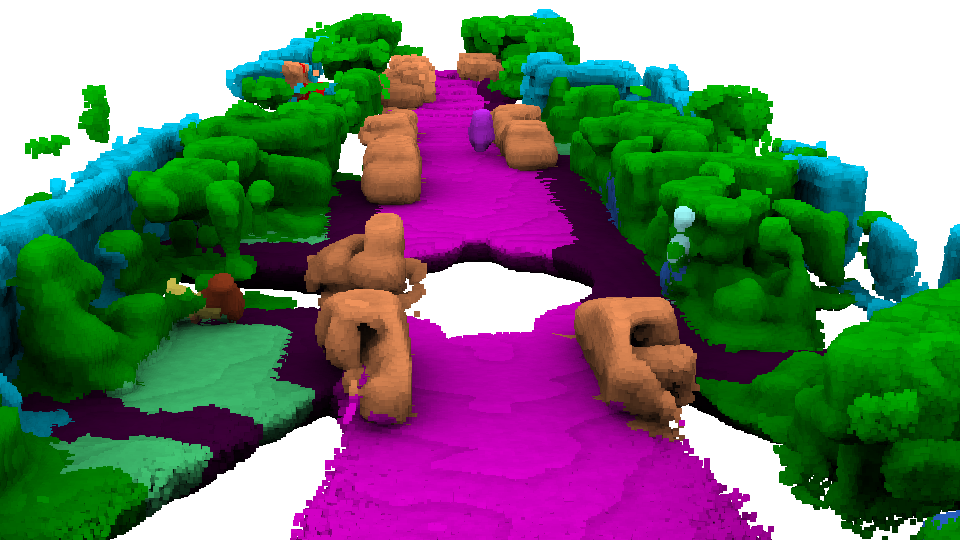} & 
        \includegraphics[width=0.245\linewidth]{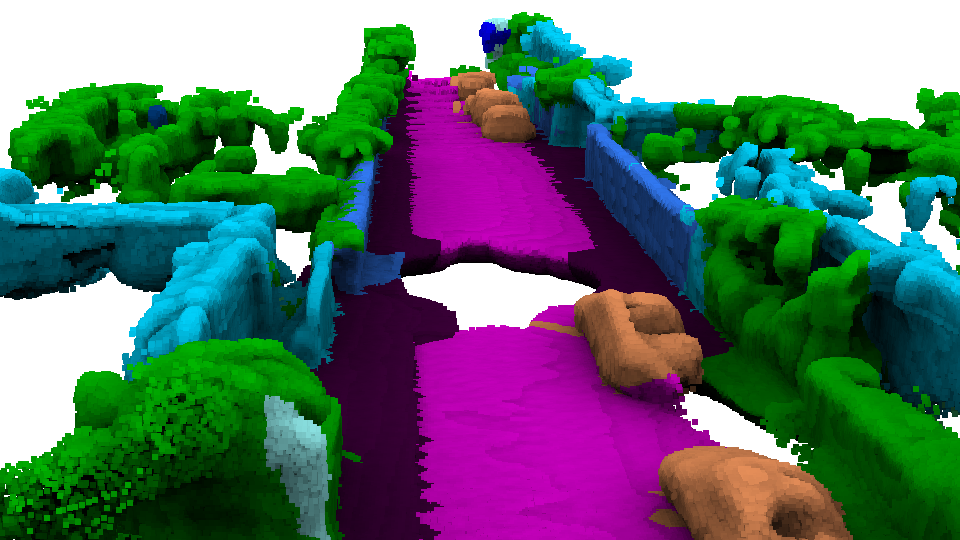} \\

        \multicolumn{4}{c}{(a) SemanticKITTI}\\
        \\

        \includegraphics[width=0.245\linewidth]{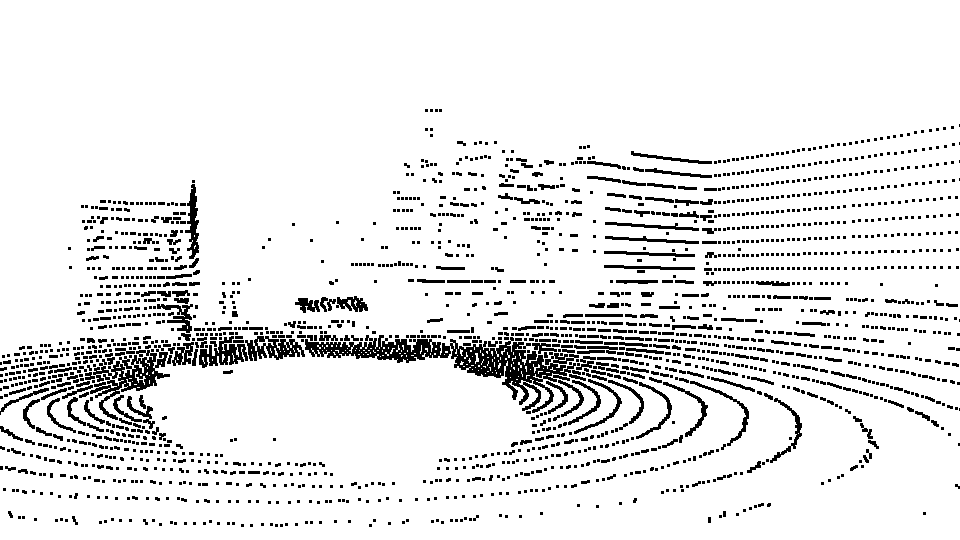} &
        \includegraphics[width=0.245\linewidth]{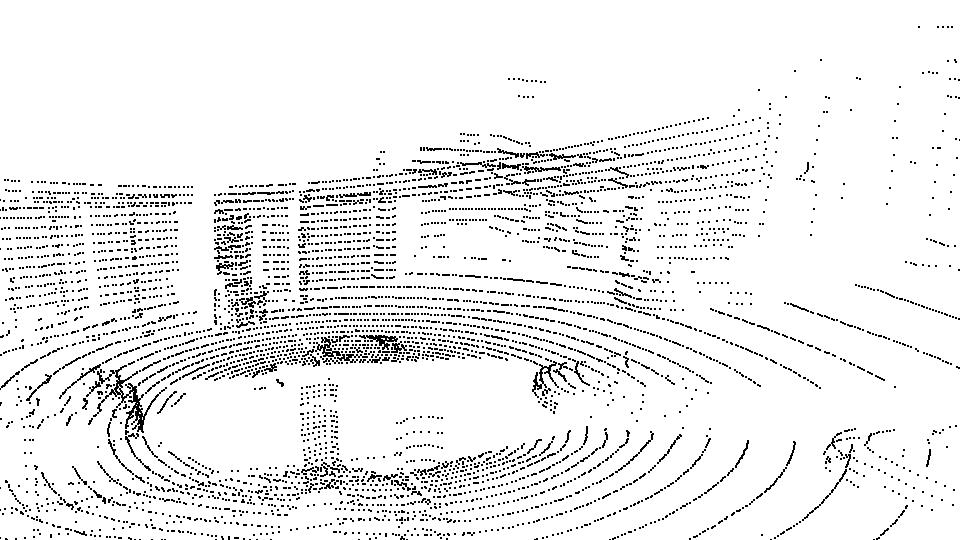} & 

        \includegraphics[width=0.245\linewidth]{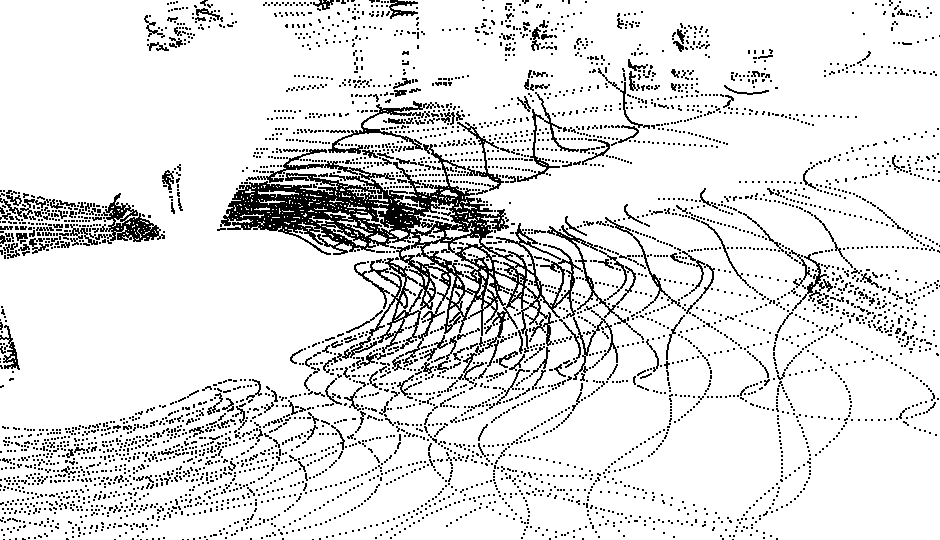} &
        \includegraphics[width=0.245\linewidth]{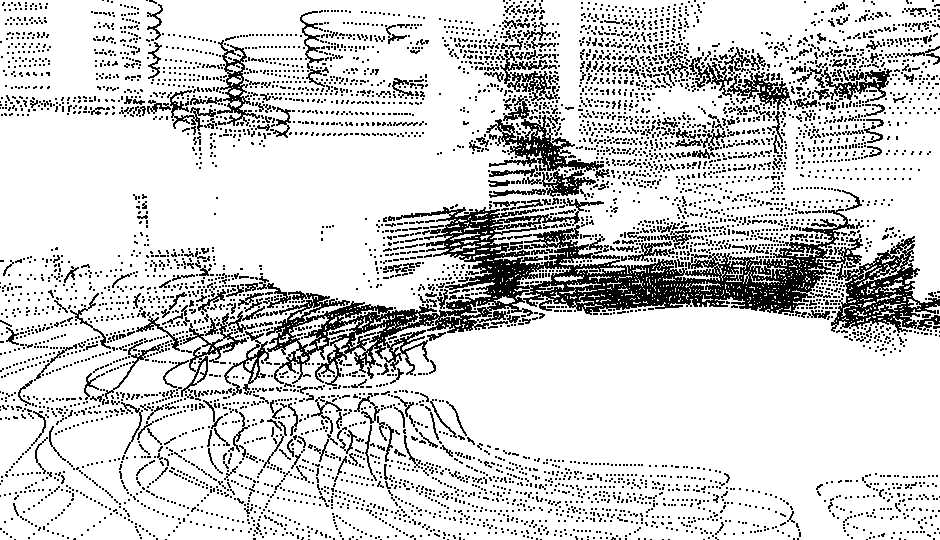} \\
        
        \includegraphics[width=0.245\linewidth]{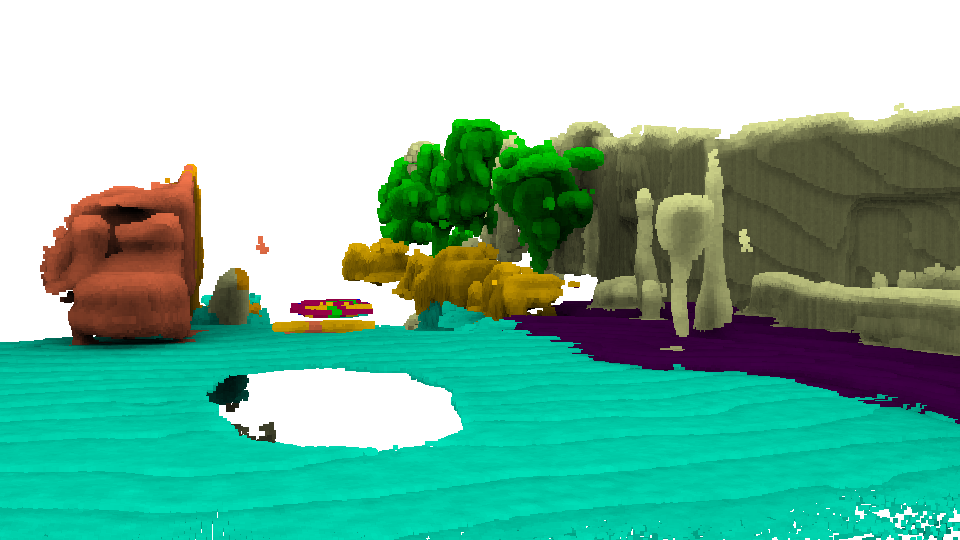} &
        \includegraphics[width=0.245\linewidth]{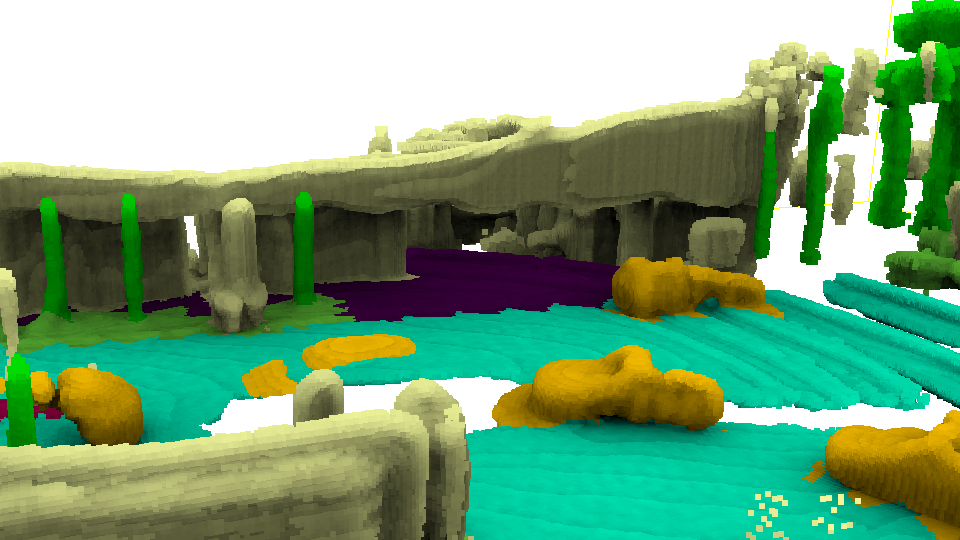} &

        \includegraphics[width=0.245\linewidth]{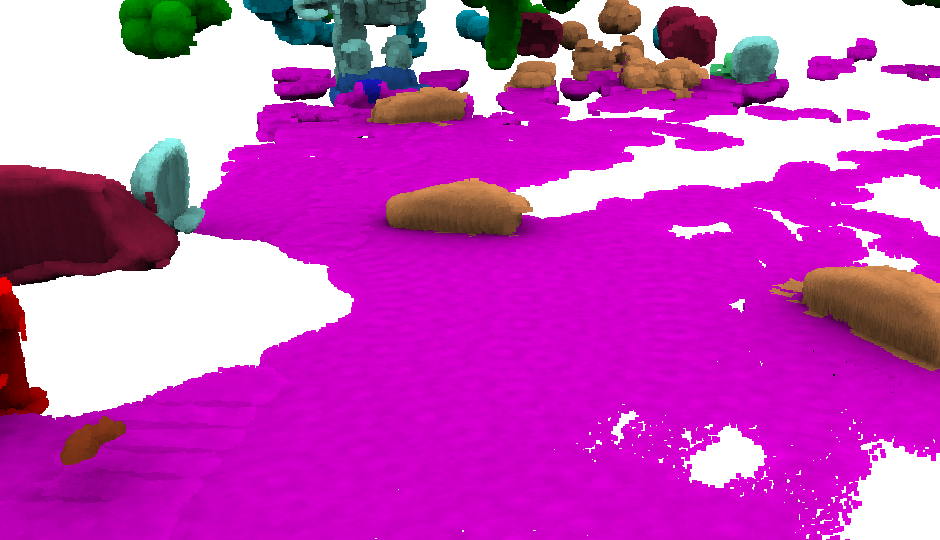} &
        \includegraphics[width=0.245\linewidth]{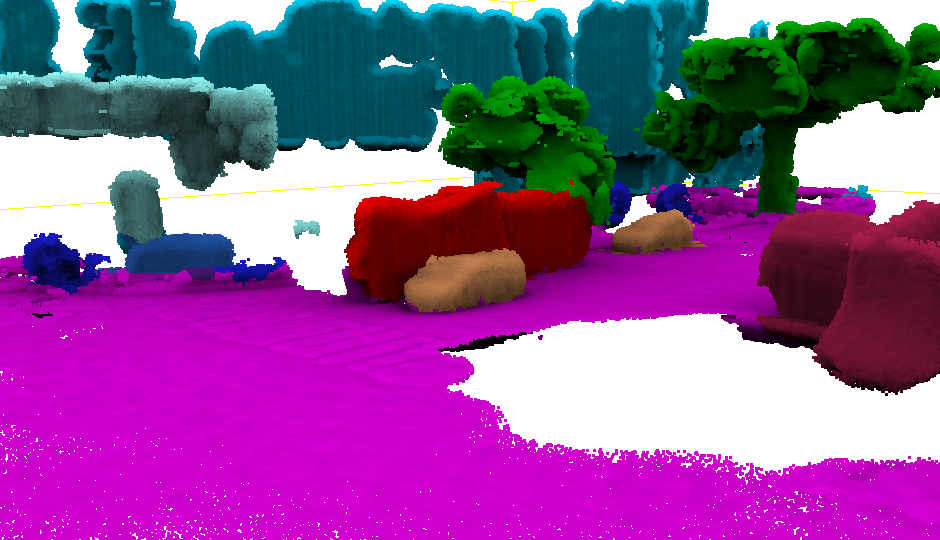} \\

        \multicolumn{2}{c|}{(b) nuScenes} &
        \multicolumn{2}{c}{(c) LivoxSimu}

        \end{tabular}

    \caption{\renaud{Semantic segmentation} predictions and occupancies on various datasets: (a)~SemanticKITTI, (b)~nuScenes, (c)~LivoxSimu.}
    \label{fig:supp_qualitative_occ_seg}
\end{figure*}

%% file: tables/table_supp_ablations.tex
\begin{table*}
\centering
\small
\begin{tabular}{cccc|ccccc|c}
\toprule
Input     & Loss              & Search & Num.     & \multicolumn{5}{c|}{Runs} & Average \\
Intensity & $\mathcal{L}_I$   & radius (m) & epochs   & \multicolumn{5}{c|}{} & and std dev\\
\midrule
\xmark & \xmark & 1.0 & 100 & 36.45 & \bf36.68 & 35.94 & 36.38 & 36.64 & 36.42 $\pm$0.30 \\
\cmark & \xmark & 1.0 & 100 & \bf38.38 & 38.17 & 38.12 & 38.23 & 38.14 & 38.21 $\pm$0.10 \\
\cmark & \cmark & 1.0 & 100 & 38.48 & 38.54 & 38.31 & \bf38.85 & 38.01 & \bf38.44 $\pm$0.31 \\
\midrule
\cmark & \cmark & 0.5 & 100 & 37.60 & \bf37.77 & 37.74 & 37.25 & 37.60 & 37.59 $\pm$0.21 \\
\cmark & \cmark & 1.0 & 100 & 38.48 & 38.54 & 38.31 & \bf38.85 & 38.01 & \bf38.44 $\pm$0.31 \\
\cmark & \cmark & 2.0 & 100 & 38.06 & 37.93 & \bf38.68 & 38.45 & 37.68 & 38.16 $\pm$0.40 \\
\cmark & \cmark & 4.0 & 100 & 36.18 & 36.89 & 36.40 & 36.38 & 35.88 & 36.35 $\pm$0.37 \\
\bottomrule
\end{tabular}
    
\caption{Ablation study on nuScenes custom ablation validation set, 1\%. Details of experiments.}
\label{tab:appendix_ablations}
\end{table*}

%% file: tables/table_supp_nuscenes.tex
\begin{table*}

\centering
\small

\begin{tabular}{cll|ccccc|c}
\toprule
\% & Backbone & Method          & \multicolumn{5}{c|}{Runs} & Average and std \\
\midrule
0.1\% & MinkUNet &No pre-training         & 21.88 & 21.21 & \bf22.05 & 21.08 & 21.96 & 21.64 $\pm$0.45\\
&&PointContrast~\cite{pointcontrast}   & 26.39	& 27.35	& \bf27.82 & 26.95 & 26.89 & \rone{27.08 $\pm$0.54} \\
&&DepthContrast~\cite{depthcontrast}   & \bf21.89 & 21.88 & 21.63 & 21.87 & 21.27 & \rthree{21.71 $\pm$0.27}\\
&&\method{}       & 26.62 & \bf26.86 & 25.99 & 25.59 & 26.08 & \rtwo{26.23 $\pm$0.51} \\
\cmidrule{2-9}
&SPVCNN & No pre-training & 21.97 & 22.30 & 22.09 & 22.18 & \bf22.45 & 22.20 $\pm$0.19\\
&& \method{} & 24.40 & 24.16 & \bf25.86 & 25.73 & 23.93 & \rone{24.82 $\pm$0.91}\\

\midrule\midrule
1\% & MinkUNet & No pre-training		& 34.86	& 35.09	& 34.72	& 34.72	& \bf35.55	& \rthree{34.99 $\pm$0.35}\\
&& PointContrast~\cite{pointcontrast}	& 37.24	& 37.24	& 36.25	& 36.76	& \bf37.36	& \rtwo{36.97 $\pm$0.46}\\
&& DepthContrast~\cite{depthcontrast}	& 34.51	& 34.74	& \bf35.38	& 34.23	& 34.07	& 34.59 $\pm$0.51\\
&& \method{}                            & 37.42	& 37.52	& 37.15	& 37.11	& \bf37.94	& \rone{37.43 $\pm$0.34}\\
\cmidrule{2-9}
&SPVCNN  & No pre-training              & 34.27	& \bf34.94	& 34.26	& 34.10	& 34.37	& 34.39 $\pm$0.32\\
        && \method{}                    & 37.24	& 37.14	& 37.55	& 37.24	& \bf37.74	& \rone{37.38 $\pm$0.25}\\

\midrule\midrule
10\% & MinkUNet & No pre-training		& 57.62	& \bf57.66	& 57.31	& 56.70	& 57.19	& 57.30 $\pm$0.39\\
&&PointContrast~\cite{pointcontrast}	& 59.00	& 58.73	& 58.66	& 58.96	& \bf59.05	& \rtwo{58.88 $\pm$0.17}\\
&&DepthContrast~\cite{depthcontrast}	& \bf58.03	& 57.00	& 57.36	& 57.56	& 56.90	& \rthree{57.37 $\pm$0.46}\\
&&\method{}                             & 58.63	& 58.62	& 59.11	& 59.28	& \bf59.35	& \rone{59.00 $\pm$0.35}\\
\cmidrule{2-9}
&SPVCNN & No pre-training               & \bf57.37	& 56.97	& 57.34	& 56.75	& 57.18	& 57.12 $\pm$0.26\\
&& \method{}                            & 58.15	& 58.56	& 58.42	& 58.48	& \bf58.60	& \rone{58.44 $\pm$0.18}\\

\midrule\midrule
50\% & MinkUNet & No pre-training		& 68.80	& 68.90	& 68.94	& \bf69.31	& 69.01	& 68.99 $\pm$0.19\\
&&PointContrast~\cite{pointcontrast}	& 69.15	& 69.09	& 69.39	& 69.42	& \bf69.75	& \rtwo{69.36 $\pm$0.26}\\
&&DepthContrast~\cite{depthcontrast}	& 69.12	& 69.04	& 69.38	& \bf69.57	& 68.66	& \rthree{69.15 $\pm$0.35}\\
&&\method{}                             & 69.69	& 69.58	& 69.93	& 69.66	& \bf70.17	& \rone{69.81 $\pm$0.24}\\
\cmidrule{2-9}
&SPVCNN & No pre-training               & \bf69.24	& 69.06	& 68.68	& 68.74	& 69.09	& 68.96 $\pm$0.24\\
&& \method{}                            & 69.55	& \bf69.77	& 69.47	& 69.24	& 69.65	& \rone{69.54 $\pm$0.20}\\

\midrule\midrule
100 \% & MinkUNet & No pre-training     & 71.21	& \bf71.35	& 71.20	& 70.93	& 71.32	& \rtwo{71.20 $\pm$0.17}\\
&&PointContrast~\cite{pointcontrast}    & 71.12	& 71.27	& 70.90	& 70.94	& \bf71.31	& 71.11 $\pm$0.19\\
&&DepthContrast~\cite{depthcontrast}    & 71.31	& 71.20	& 71.30	& 70.81	& \bf71.36	& \rtwo{71.20 $\pm$0.22}\\
&&\method{}                             & \bf71.95	& 71.92	& 71.60	& 71.88	& 71.39	& \rone{71.75 $\pm$0.24}\\
\cmidrule{2-9}
&SPVCNN & No pre-training               & 70.82	& 70.79	& 70.56	& \bf70.86 & 70.41  & 70.69 $\pm$0.19\\
&& \method{}                            & 71.41	& 71.18	& 70.99	& 71.20	& \bf71.48	& \rone{71.25 $\pm$0.20}\\
\bottomrule
\end{tabular}

\caption{NuScenes. Details of experiments.}
\label{tab:appendix_nuscenes}
\end{table*}

%% file: tables/table_supp_semantickitti.tex
\begin{table*}
\small
\centering

\begin{tabular}{cll|c|ccccc|c}
\toprule
\% & Backbone & Method          & from~\cite{segcontrast}   & \multicolumn{5}{c|}{Runs} & Average and std \\
\midrule
0.1\% & MinkUNet &No pre-training         & \cellcolor{black!10}25.59 & \bf30.22 & 29.99 & 29.74 & 30.15 & 29.77 & 29.97 $\pm$0.22\\
&&PointContrast~\cite{pointcontrast}   & \cellcolor{black!10}28.52 & 32.79	& 31.84	& 31.88	& \bf32.96 & 32.60 & \rthree{32.41 $\pm$0.52}\\
&&DepthContrast~\cite{depthcontrast}   & \cellcolor{black!10}33.51 & 32.43	& 32.09	& \bf33.01 & 32.24 & 32.80 & \rtwo{32.51 $\pm$0.38}\\
&&SegContrast~\cite{segcontrast}     & \cellcolor{black!10}34.78 & \bf32.65 & 32.38 & 32.48 & 32.18 & 31.83 & 32.30	$\pm$0.31\\
&&\method{}       & \cellcolor{black!10}N/A   & \bf34.97 & 34.83 & 34.81 & 35.10 & 35.11 & \rone{34.96 $\pm$0.14}\\
\cmidrule{2-10}
&SPVCNN &No pre-training & \cellcolor{black!10}N/A   & \bf30.94 & 30.81 & 30.66 & 30.47 & 30.81 & 30.74 $\pm$0.18 \\
&&\method{} & \cellcolor{black!10}N/A   & 35.35 & 34.78 & 34.71 & 34.93 & \bf35.43 & \rone{35.04 $\pm$0.33} \\

\midrule\midrule
1\% & MinkUNet &No pre-training         & \cellcolor{black!10}41.70 & 45.1	& 46.32	& 46.59	& \bf46.68	& 46.49 & 46.24	$\pm$0.65 \\
&&PointContrast~\cite{pointcontrast}   & \cellcolor{black!10}43.40	& 47.71	& 47.97	& 48.22	& 47.25	& \bf48.43 & 47.92	$\pm$0.46 \\
&&DepthContrast~\cite{depthcontrast}   & \cellcolor{black!10}46.41	& \bf49.62	& 49.12	& 48.59	& 48.94	& 48.74 & \rtwo{49.00 $\pm$0.40} \\
&&SegContrast~\cite{segcontrast}     & \cellcolor{black!10}47.41	& 48.91	& \bf49.35	& 48.87	& 48.81	& 48.54 & \rthree{48.90 $\pm$0.29} \\
&&\method{}       & \cellcolor{black!10}N/A   & 50.04	& 50.28	& 49.43	& \bf50.41	& 50.02	& \rone{50.04 $\pm$0.38} \\
\cmidrule{2-10}
&SPVCNN &No pre-training & \cellcolor{black!10}N/A   &46.24 & \bf47.23 & 46.53 & 46.26 & 46.67 & 46.59 $\pm$0.40\\
&&\method{} & \cellcolor{black!10}N/A   & 49.34 & \bf49.75 & 49.05 & 48.90 & 48.32 &  \rone{49.07 $\pm$0.53} \\

\midrule\midrule
10\% & MinkUNet &No pre-training         & \cellcolor{black!10}53.87 & 57.04 & \bf58.74 & 57.71 & 56.27 & 58.01 & 57.55 $\pm$0.94\\
&&PointContrast~\cite{pointcontrast}   & \cellcolor{black!10}53.79 & 59.48 & 59.78 & \bf60.44 & 59.26 & 59.56 & \rthree{59.70 $\pm$0.45}\\
&&DepthContrast~\cite{depthcontrast}   & \cellcolor{black!10}56.29 & 59.49 & 60.74 & 60.27 & 60.46 & \bf60.75 & \rtwo{60.34 $\pm$0.52}\\
&&SegContrast~\cite{segcontrast}     & \cellcolor{black!10}55.21 & \bf59.63 & 58.57 & 58.45 & 58.78 & 58.29 & 58.74 $\pm$0.53\\
&&\method{}       & \cellcolor{black!10}N/A & 60.41 & 60.45 & 60.47 & \bf60.54 & 60.43 & \rone{60.46 $\pm$0.05}\\
\cmidrule{2-10}
&SPVCNN &No pre-training & \cellcolor{black!10}N/A   &58.8 & 58.95 & 59.21 & \bf59.47 & 57.85 & 58.86 $\pm$0.62 \\
&&\method{} & \cellcolor{black!10}N/A   & 60.71 & 60.32 & \bf60.97 & 60.32 & 60.66 & \rone{60.60 $\pm$0.28} \\

\midrule\midrule
50\% & MinkUNet &No pre-training         & \cellcolor{black!10}58.34 & 61.48 & \bf62.33 & 61.88 & 61.80 & 61.31 & 61.76 $\pm$0.39\\
&&PointContrast~\cite{pointcontrast}   & \cellcolor{black!10}57.30 & 62.68 & 62.91 & 62.54 & 62.35 & \bf63.19	& \rthree{62.73 $\pm$0.33}\\
&&DepthContrast~\cite{depthcontrast}   & \cellcolor{black!10}58.54 & 63.24 & \bf63.31 & 63.16 & 62.44 & 62.37	& \rtwo{62.90 $\pm$0.46}\\
&&SegContrast~\cite{segcontrast}     & \cellcolor{black!10}58.33 & \bf62.58 & 62.20 & 61.61 & 61.74 & 62.46 & 62.12 $\pm$0.43\\
&&\method{}       & \cellcolor{black!10}N/A   & 63.09 & 63.43 & 62.99 & 63.28 & \bf64.15 & \rone{63.39 $\pm$0.46}\\
\cmidrule{2-10}
&SPVCNN &No pre-training & \cellcolor{black!10}N/A   &61.32 & 62.15 & 61.61 & 61.7 & \bf62.39 & 61.83 $\pm$0.43 \\
&&\method{} & \cellcolor{black!10}N/A   & 63.4 & 63.4 & 63.45 & \bf64.08 & 63.44 & \rone{63.55 $\pm$0.29} \\

\midrule\midrule
100\% & MinkUNet &No pre-training         & \cellcolor{black!10}59.63 & 62.49 & 62.35 & 62.98 & 62.50 & \bf63.06	& 62.68 $\pm$0.32\\
&&PointContrast~\cite{pointcontrast}   & \cellcolor{black!10}59.77 & 63.57 & 63.14 & 63.13 & \bf63.95 & 63.26 & \rthree{63.41 $\pm$0.35}\\
&&DepthContrast~\cite{depthcontrast}   & \cellcolor{black!10}59.88 & 63.76 & \bf64.31 & 63.52 & 63.54 & 64.12 & \rone{63.85 $\pm$0.35}\\
&&SegContrast~\cite{segcontrast}     & \cellcolor{black!10}60.53 & \bf62.64 & 61.57 & 62.53 & 62.24 & 62.45 & 62.29 $\pm$0.43\\
&&\method{}       & \cellcolor{black!10}N/A   & \bf{64.29} & 63.75 & 63.75 & 63.34 & 63.07 & \rtwo{63.64 $\pm$0.46}\\
\cmidrule{2-10}
&SPVCNN &No pre-training & \cellcolor{black!10}N/A   &62.39 & 62.86 & 62.33 & \bf62.88 & 62.82 & 62.66 $\pm$0.27 \\
&&\method{} & \cellcolor{black!10}N/A   & 63.60 & \bf64.04 & 63.59 & 63.93 & 63.76 & \rone{63.78 $\pm$0.20} \\
\bottomrule
\end{tabular}

\caption{SemanticKITTI. Details of experiments.}
\label{tab:appendix_semantickitti}
\end{table*}

%% file: tables/table_supp_semanticposs.tex
\begin{table*}
\centering
\small

\begin{tabular}{cll|ccccc|c}
\toprule
\% & Backbone & Method          & \multicolumn{5}{c|}{Runs} & Average and std \\
\midrule
0.1\% & MinkUNet &No pre-training         & 37.23 & \bf37.53 & 36.37 & 36.72 & 36.59 & 36.89 $\pm$0.48 \\
&&PointContrast~\cite{pointcontrast}   & 39.22 & \bf40.60 & 38.56 & 39.24 & 38.73 & 39.27 $\pm$0.80\\
&&DepthContrast~\cite{depthcontrast}   & 38.69 & 39.35 & \bf41.16 & 39.87 & 39.25 & \rthree{39.66 $\pm$0.94}\\
&&SegContrast~\cite{segcontrast}     & 41.72	& \bf42.89 & 41.45 & 41.74 & 40.68 & \rone{41.70 $\pm$0.79}\\
&&\method{}       & 40.04	& 41.23 & \bf41.29 & 40.88 & 39.83 & \rtwo{40.65 $\pm$0.68}\\

\midrule\midrule
1\% & MinkUNet &No pre-training			& \bf46.99 & 46.23 & 46.09 & 46.33 & 46.47 & 46.42	$\pm$0.35 \\
&&PointContrast~\cite{pointcontrast}	& \bf48.45 & 48.26 & 48.40 & 48.43 & 47.11 & 48.13	$\pm$0.58\\
&&DepthContrast~\cite{depthcontrast}	& 48.08 & 48.78 & 48.29 & 48.36 & \bf48.96 & \rthree{48.49	$\pm$0.36}\\
&&SegContrast~\cite{segcontrast}		& 48.94 & \bf50.02 & 49.34 & 49.08 & 49.66 & \rtwo{49.41	$\pm$0.44}\\
&&\method{}       & \bf50.55 & 48.85 & 49.02 & 49.86 & 49.51 & \rone{49.56	$\pm$0.68}\\

\midrule\midrule
10\% & MinkUNet &No pre-training			& 54.29 & 53.96 & 54.53 & \bf54.95 & 54.60 & 54.47 $\pm$0.37\\
&&PointContrast~\cite{pointcontrast}	& \bf55.91 & 54.60 & 54.77 & 54.96 & 55.29 & 55.11	$\pm$0.52\\
&&DepthContrast~\cite{depthcontrast}	& \bf56.17 & 55.81 & 55.17 & 55.82 & 55.96 & \rtwo{55.79 $\pm$0.37}\\
&&SegContrast~\cite{segcontrast}		& 55.48 & 55.41 & 55.39 & \bf55.50 & 54.97 & \rthree{55.35 $\pm$0.22}\\
&&\method{}       & 56.16 & \bf56.19 & 55.43 & 55.15 & 56.14 & \rone{55.81 $\pm$0.49}\\

\midrule\midrule
50\% & MinkUNet &No pre-training			& \bf55.48 & 55.19 & 55.40 & 55.27 & 55.04 & 55.28 $\pm$0.17\\
&&PointContrast~\cite{pointcontrast}	& 55.84 & 56.46 & 55.62 & 55.90 & \bf57.01 & \rthree{56.17 $\pm$0.56}\\
&&DepthContrast~\cite{depthcontrast}	& 54.67 & 56.35 & 56.12 & 56.26 & \bf56.73 & 56.03 $\pm$0.79\\
&&SegContrast~\cite{segcontrast}		& 56.84 & 55.89 & 54.82 & \bf56.89 & 56.76 & \rtwo{56.24 $\pm$0.89}\\
&&\method{}       & \bf57.07 & 55.56 & 56.71 & 56.13 & 56.30 & \rone{56.35 $\pm$0.58}\\

\midrule\midrule
100\% & MinkUNet &No pre-training         & 54.52 & \bf55.52 & \bf55.52 & 55.10 & 54.83 & 55.10 $\pm$0.44\\
&&PointContrast~\cite{pointcontrast}   & 55.80 & 56.02 & 55.48 & \bf57.13 & 56.40 & 56.17 $\pm$0.63\\
&&DepthContrast~\cite{depthcontrast}   & \bf57.38 & 56.36 & 56.40 & 56.29 & 56.00 & \rtwo{56.49 $\pm$0.52}\\
&&SegContrast~\cite{segcontrast}     & 56.24 & 56.46 & 56.22 & \bf57.17 & 55.83 & \rthree{56.38 $\pm$0.49}\\
&&\method{}       & 56.88 & \bf58.23 & 55.45 & 56.01 & 56.88 & \rone{56.69 $\pm$1.05} \\
\bottomrule
\end{tabular}

\caption{SemanticPOSS. Details of experiments.}
\label{tab:appendix_semanticposs}
\end{table*}

%% file: tables/table_supp_livox.tex
\begin{table*}
\centering
\small

\begin{tabular}{cll|ccccc|c}
\toprule
\% & Backbone & Method          & \multicolumn{5}{c|}{Runs} & Average and std \\
\midrule
0.1\% & MinkUNet &No pre-training & 47.43 & \bf48.38 & 48.03 & 48.21 & 48.10 & 48.03 $\pm$0.36\\
&&\method{}       & 51.47 & 52.31 & \bf54.35 & 52.28 & 52.45 & \rone{52.57 $\pm$1.07}\\

\midrule\midrule
1\% & MinkUNet &No pre-training	& 63.73 & 63.33 & 63.46 & \bf64.38 & 64.03 & 63.79 $\pm$0.43\\
&&\method{}       & \bf65.86 & 65.48 & 65.16 & 65.50 & 65.24 & \rone{65.45	$\pm$0.27}\\

\midrule\midrule
10\% & MinkUNet &No pre-training	& 66.51 & \bf66.84 & 66.67 & 66.60 & 66.64 & 66.65 $\pm$0.12\\
&&\method{}       & 67.30 & 68.08 & 67.43 & \bf68.10 & 67.82 & \rone{67.75 $\pm$0.37}\\

\midrule\midrule
50\% & MinkUNet &No pre-training	& 68.35 & \bf68.87 & 68.10 & 68.60 & 68.52 & 68.49 $\pm$0.29\\
&&\method{}       & 69.55 & 69.41 & 69.66 & 69.55 & \bf69.73 & \rone{69.58 $\pm$0.12}\\

\midrule\midrule
100\% & MinkUNet &No pre-training & 68.91 & 68.87 & 68.82 & \bf69.25 & 68.68 & 68.91 $\pm$0.21\\
&&\method{}       & 69.37 & 69.86 & 69.62 & \bf70.14 & 69.61 & \rone{69.72 $\pm$0.29}\\
\bottomrule
\end{tabular}

\caption{Livox Synthetic Dataset. Details of experiments.}
\label{tab:appendix_livox}
\end{table*}

%% file: tables/table_supp_kitti3d.tex
\begin{table*}
\small
\centering

\setlength{\tabcolsep}{4pt}

\begin{tabular}{llll|ccc|ccc|ccc}
\toprule
Backbone & Metric & Method &Pre-training & \multicolumn{3}{c|}{Car} & \multicolumn{3}{c|}{Pedestrian} & \multicolumn{3}{c}{Cyclist} \\
& set & & Easy & Mod. & Hard &Easy& Mod. & Hard & Easy& Mod. & Hard\\
\midrule
SECOND & 2D object detection 
& Scratch$^\dagger$ & -             & 95.84 & 94.49 & 92.00 & 68.27 & 64.69 & 61.15 & 91.02 & 78.88 & 76.00 \\
\greyrule{\cmidrule{3-13}}
&& \method{}        & KITTI         & \ptwo{96.88} & \mone{94.43} & \mone{91.89} & \pthree{70.44} & \pthree{67.60} & \pfour{64.40} & \ptwo{91.67} & \pthree{81.67} & \pthree{78.01} \\
&&                  & KITTI-360     & \mone{95.69} & \mone{94.34} & \mone{91.81} & \ptwo{69.76} & \pthree{67.12} & \pthree{64.12} & \pthree{93.26} & \pthree{80.97} & \ptwo{77.87} \\
&&                  & nuScenes      & \ptwo{97.48} & \pone{94.70} & \ptwo{93.56} & \pfive{72.39} & \pthree{68.69} & \pfive{65.86} & \pone{91.34} & \pthree{81.64} & \pthree{78.46} \\
\cmidrule{2-13}
&Bird's eye view 
& Scratch$^\dagger$ & -             & 93.76 & 89.82 & 87.65 & 59.74 & 54.85 & 50.56 & 87.19 & 70.96 & 68.00 \\
\greyrule{\cmidrule{3-13}}
&& \method{}        & KITTI         & \pone{93.82} & \pone{90.16} & \pone{88.00} & \pone{60.38} & \ptwo{56.01} & \pthree{52.58} & \pone{87.90} & \pthree{73.39} & \ptwo{69.15} \\
&&                  & KITTI-360     & \mtwo{92.38} & \mone{89.60} & \pone{87.79} & \pthree{61.75} & \pthree{57.36} & \pfour{53.91} & \pthree{89.26} & \pthree{73.74} & \ptwo{69.48} \\
&&                  & nuScenes      & \pone{94.64} & \pone{90.77} & \pone{88.24} & \pfive{64.32} & \pfive{59.13} & \pfive{55.25} & \mone{86.92} & \pfour{74.58} & \pthree{70.17} \\
\cmidrule{2-13}
&3D object detection
& Scratch$^\dagger$ & -            & 90.20 & 81.50 & 78.61 & 53.89 & 48.82 & 44.56 & 82.59 & 65.72 & 62.99 \\
\greyrule{\cmidrule{3-13}}
&& \method{}        & KITTI        & \pone{90.76} & \pone{81.97} & \pone{79.10} & \pthree{56.30} & \pfour{51.93} & \pfour{48.03} & \ptwo{83.71} & \pfour{69.14} & \pthree{65.27} \\
&&                  & KITTI-360    & \mtwo{88.95} & \pone{81.79} & \pone{78.92} & \pfour{57.83} & \pfour{52.45} & \pfour{48.32} & \pfive{86.76} & \pfive{70.68} & \pfour{66.56} \\
&&                  & nuScenes     & \pone{90.21} & \pone{81.78} & \pone{78.97} & \psix{59.56} & \psix{54.24} & \psix{50.27} & \mtwo{81.12} & \pthree{68.19} & \ptwo{64.10} \\
\cmidrule{2-13}
&Orientation similarity
& Scratch$^\dagger$ & -            & 95.83 & 94.35 & 91.79 & 63.70 & 59.52 & 55.85 & 90.86 & 78.44 & 75.52 \\
\greyrule{\cmidrule{3-13}}
&& \method{}        & KITTI        & \ptwo{96.86} & \mone{94.30} & \mone{91.65} & \pthree{66.44} & \pfour{62.82} & \pfour{59.17} & \ptwo{91.36} & \pthree{81.07} & \ptwo{77.35} \\
&&                  & KITTI-360    & \mone{95.68} & \mone{94.24} & \pone{91.63} & \ptwo{65.58} & \pthree{62.25} & \pthree{58.52} & \pthree{92.35} & \pone{78.70} & \pone{75.64} \\
&&                  & nuScenes     & \ptwo{97.45} & \pone{94.54} & \ptwo{93.30} & \pfour{67.67} & \pfour{63.33} & \pfive{60.28} & \pone{91.01} & \pthree{80.75} & \ptwo{77.49} \\
\midrule\midrule
PV-RCNN & 2D object detection
& Scratch$^\dagger$ & -            & 97.86 & 94.39 & 93.92 & 73.84 & 68.68 & 65.53 & 94.34 & 81.89 & 77.36 \\
\greyrule{\cmidrule{3-13}}
&& \method{}        & KITTI        & \pone{98.26} & \pone{94.42} & \pone{94.07} & \pthree{76.05} & \pthree{70.89} & \ptwo{67.50} & \pone{95.32} & \ptwo{83.41} & \pfour{80.42}\\
&&                  & KITTI-360    & \pone{98.04} & \pone{94.42} & \pone{94.11} & \pthree{76.75} & \pthree{71.15} & \pone{67.40} & \pone{95.18} & \ptwo{83.59} & \ptwo{78.70}\\
&&                  & nuScenes     & \mtwo{96.12} & \pone{94.45} & \pone{93.99} & \mone{73.70} & \pone{68.70} & \mone{65.31} & \pone{94.61} & \mone{81.86} & \ptwo{78.67}\\
\cmidrule{2-13}
&Bird's eye view
& Scratch$^\dagger$ & -            & 94.65 & 90.61 & 88.56 & 68.28 & 60.62 & 55.95 & 92.52 & 75.03 & 70.40 \\
\greyrule{\cmidrule{3-13}}
&& \method{}        & KITTI        & \pone{94.82} & \pone{90.75} & \pone{88.67} & \pone{68.93} & \ptwo{61.88} & \ptwo{57.74} & \pone{93.18} & \pthree{77.73} & \pthree{73.09}\\
&&                  & KITTI-360    & \mone{94.40} & \mone{90.60} & 88.56 & \pfour{72.04} & \pthree{63.40} & \pfour{59.05} & \pthree{95.11} & \pthree{77.25} & \pthree{73.37}\\
&&                  & nuScenes     & \mtwo{93.10} & \pone{90.64} & \mone{88.53} & \pone{68.72} & \pone{60.92} & \ptwo{56.96} & \pone{93.11} & \ptwo{76.74} & \pthree{73.06}\\
\cmidrule{2-13}
& 3D object detection
& Scratch$^\dagger$ & -            & 91.74 & 84.60 & 82.29 & 65.51 & 57.49 & 52.71 & 91.37 & 71.51 & 66.98 \\
\greyrule{\cmidrule{3-13}}
&& \method{}        & KITTI        & \pone{91.90} & \pone{84.72} & \pone{82.55} & \pone{65.57} & \ptwo{58.49} & \ptwo{53.75} & \ptwo{92.52} & \pfour{75.06} & \pfour{70.48}\\
&&                  & KITTI-360    & \pone{92.13} & \pone{84.68} & \pone{82.58} & \pfour{68.72} & \pthree{60.16} & \pthree{54.87} & \ptwo{92.86} & \pthree{74.04} & \pthree{69.30}\\
&&                  & nuScenes     & \pone{92.31} & \pone{84.86} & \pone{82.61} & \pone{65.60} & \pone{57.76} & \pone{52.96} & \pone{91.70} & \pfour{74.98} & \pfour{70.67}\\
\cmidrule{2-13}
& Orientation similarity
& Scratch$^\dagger$ & -            & 97.84 & 94.25 & 93.70 & 69.73 & 63.89 & 60.31 & 94.20 & 81.00 & 76.47 \\
\greyrule{\cmidrule{3-13}}
&& \method{}        & KITTI        & \pone{98.23} & \pone{94.31} & \pone{93.89} & \pone{70.07} & \ptwo{64.94} & \pone{61.09} & \pone{95.15} & \ptwo{82.84} & \pfour{79.79}\\
&&                  & KITTI-360    & \pone{98.02} & \pone{94.32} & \pone{93.91} & \pthree{72.55} & \pthree{66.62} & \pthree{62.64} & \pone{94.85} & \pthree{83.07} & \ptwo{78.14}\\
&&                  & nuScenes     & \mtwo{96.09} & \pone{94.29} & \pone{93.76} & \mthree{67.66} & \mtwo{62.74} & \mone{59.37} & \pone{94.23} & \pone{81.07} & \ptwo{77.86}\\
\bottomrule
\multicolumn{13}{l}{$^\dagger$: retrained by ourselves, scores may vary from main paper.}
\end{tabular}

Color scale:
\colorbox{MyRed!25}{[-3,-2[}
\colorbox{MyRed!15}{[-2,-1[}
\colorbox{MyRed!5}{[-1,0[}
\colorbox{MyGreen!5}{[0,1[}
\colorbox{MyGreen!15}{[1,2[}
\colorbox{MyGreen!25}{[2,3[}
\colorbox{MyGreen!35}{[3,4[}
\colorbox{MyGreen!45}{[5,6[}
\colorbox{MyGreen!55}{[6,7[}

\caption{KITTI3D detection. Details of experiments. Cells are colored according to difference with from-scratch pre-training.}
\label{tab:appendix_detection_kitti3d}
\end{table*}

%% file: tables/table_supp_once.tex
\begin{table*}
\centering
\small
\setlength{\tabcolsep}{3pt}
\begin{tabular}{l|cccc|cccc|cccc|c}
\toprule
Method  & \multicolumn{4}{c|}{Vehicle} & \multicolumn{4}{c|}{Pedestrian} & \multicolumn{4}{c|}{Cyclist} & mAP \\
        & overall & 0-30 & 30-50 & 50-inf & overall & 0-30 & 30-50 & 50-inf & overall & 0-30 & 30-50 & 50-inf & \\
\midrule
\midrule
\multicolumn{14}{c}{$U_{small}$} \\
\midrule
baseline        & 71.19 & 84.04 & 63.02 & 47.25 & \rtwo{26.44} & \rtwo{29.33} & \rtwo{24.05} & \rone{18.05} & 58.04 & 69.96 & 52.43 & 34.61 & 51.89 \\
BYOL            & 68.02 & 81.01 & 60.21 & 44.17 & 19.50 & 22.16 & 16.68 & 12.06 & 50.61 & 62.46 & 44.29 & 28.18 & 46.04 \tminus{-5.85} \\
PointContrast   & 71.07 & 83.31 & 64.90 & 49.34 & 22.52 & 23.73 & 21.81 & 16.06 & 56.36 & 68.11 & 50.35 & 34.06 & 49.98 \tminus{-1.91} \\
SwAV            & \rtwo{72.71} & 83.68 & \rtwo{65.91} & \rtwo{50.10} & 25.13 & 27.77 & 22.77 & 16.36 & 58.05 & 69.99 & 52.23 & \rtwo{34.86} & 51.96 \tplus{+0.07} \\
DeepCluster     & \rone{73.19} & \rtwo{84.25} & \rone{66.86} & \rone{50.47} & 24.00 & 26.36 & 21.73 & \rtwo{16.79} & \rone{58.99} & \rone{70.80} & \rone{53.66} & \rone{36.17} & \rtwo{52.06} \tplus{+0.17} \\
\method{}       & 71.73 & \rone{84.30} & 65.21 & 48.30 & \rone{28.16} & \rone{31.45} & \rone{25.19} & 16.29 & \rtwo{58.13} & \rtwo{70.04} & \rtwo{52.76} & 33.88 & \rone{52.68} \tplus{+0.79} \\
\bottomrule

\end{tabular}

\caption{ONCE detection. Detail of experiments.}
\label{table:supp_once}
\end{table*}